\documentclass[Afour,sageh,times]{sagej}

\usepackage{moreverb,url}

\usepackage[colorlinks,bookmarksopen,bookmarksnumbered,citecolor=red,urlcolor=red]{hyperref}

\usepackage{pifont}
\usepackage{amsmath,amsfonts}
\usepackage{mathtools}
\usepackage{algorithmic}
\usepackage{algorithm}
\usepackage{array}
\usepackage{subcaption}
\usepackage{textcomp}
\usepackage{stfloats}
\usepackage{url}
\usepackage{verbatim}
\usepackage{graphicx}
\usepackage{booktabs} 
\usepackage{xcolor}
\usepackage{makecell}
\usepackage{multirow} 
\usepackage{svg}
\usepackage{bm} 
\usepackage{siunitx} 
\usepackage{ragged2e}
\usepackage[normalem]{ulem}
\newcommand{\soutMath}[1]{\ifmmode\text{\sout{\ensuremath{#1}}}\else\sout{#1}\fi}

\usepackage{amsthm,thmtools} 
\usepackage{kotex}

\newtheorem{theorem}{Theorem}
\newtheorem{lemma}{Lemma}
\newtheorem{assumption}{Assumption}
\newtheorem*{control_problem}{Geometric Robust Control}
\newtheorem*{planning_problem}{Whole-body Motion Planning}

\newcommand\blue[1]{\textcolor{black}{#1}}

\newcommand\Tanh{\text{Tanh}}

\newcommand{\cmark}{\ding{51}}
\newcommand{\xmark}{\ding{55}}

\def\volumeyear{2016}

\begin{document}

\runninghead{Lee et al.}

\title{Autonomous Aerial Manipulation at Arbitrary Pose in $\mathsf{SE}(3)$ with Robust Control and Whole-body Planning}

\author{Dongjae Lee$^*$, Byeongjun Kim$^*$, and H. Jin Kim}

\affiliation{$^*$The first two authors contributed equally to this work. \\
The authors are with the Department of Aerospace Engineering and the Automation and Systems Research Institute (ASRI), Seoul National University, Seoul 08826, South Korea}

\corrauth{H. Jin Kim, Department of Aerospace Engineering and the Automation and Systems Research Institute (ASRI), Seoul National University, Seoul 08826, South Korea.}

\email{hjinkim@snu.ac.kr}

\begin{abstract}
Aerial manipulators based on conventional multirotors can conduct manipulation only in small roll and pitch angles due to the underactuatedness of the multirotor base. If the multirotor base is capable of hovering at arbitrary orientation,
the robot can freely locate itself at any point in $\mathsf{SE}(3)$, significantly extending its manipulation workspace and enabling a manipulation task that was originally not viable.
In this work, we present a geometric robust control and whole-body motion planning framework for an omnidirectional aerial manipulator (OAM).
To maximize the strength of OAM, we first propose a geometric robust controller for a floating base. Since the motion of the robotic arm and the interaction forces during manipulation affect the stability of the floating base, the base should be capable of mitigating these adverse effects while controlling its 6D pose. We then design a two-step 
optimization-based whole-body motion planner, jointly considering the pose of the floating base and the joint angles of the robotic arm to harness the entire configuration space.
The devised two-step approach facilitates real-time applicability and enhances convergence of the optimization problem with non-convex and non-Euclidean search space. The proposed approach enables the base to be stationary at any 6D pose while autonomously carrying out sophisticated manipulation near obstacles without any collision. We demonstrate the effectiveness of the proposed framework through experiments in which an OAM performs grasping and pulling of an object in multiple scenarios, including near $90^\circ$ and even $180^\circ$ pitch angles.

\end{abstract}


\maketitle

\section{Introduction}

\subsection{Background and motivation}

Mobile manipulators with various base platforms (quadruped \cite{sleiman2021unified}, ball-balancing robot \cite{minniti2019whole}, multirotor \cite{wang2023millimeter}) have been introduced to expand the workspace of the manipulator. In particular, thanks to the capability to locate itself in wider workspace, aerial manipulators utilizing aerial robots as a base have been actively studied \cite{ollero2022past}. However, existing aerial manipulators based on conventional multirotors can conduct manipulation only in small roll, pitch angles due to the underactuatedness of the multirotor base. If additional freedom exists for a multirotor base to hover at arbitrary orientation, the workspace of the manipulator can be considerably enlarged, and such enlarged workspace may enable a manipulation task that is originally not viable. 
Although aerial manipulators based on a fully actuated multirotor base \cite{ryll20196d} are studied in \cite{tognon2019truly, nava2020direct} to provide hoverability in non-zero roll, pitch angles, there still exist non-trivial regions in orientation space (e.g. near $90^{\circ}$ pitch angle) where the multirotor cannot hover.
Several studies have been proposed to enhance versatility by serially connecting multiple aerial robots \cite{zhao2018transformable, zhao2023versatile}. However, when the task involves object grasping, these platforms require substantially large, collision-free workspace as they utilize aerial robot modules to enclose the object.

\begin{figure}
    \centering
    \includegraphics[width=1.0\linewidth]{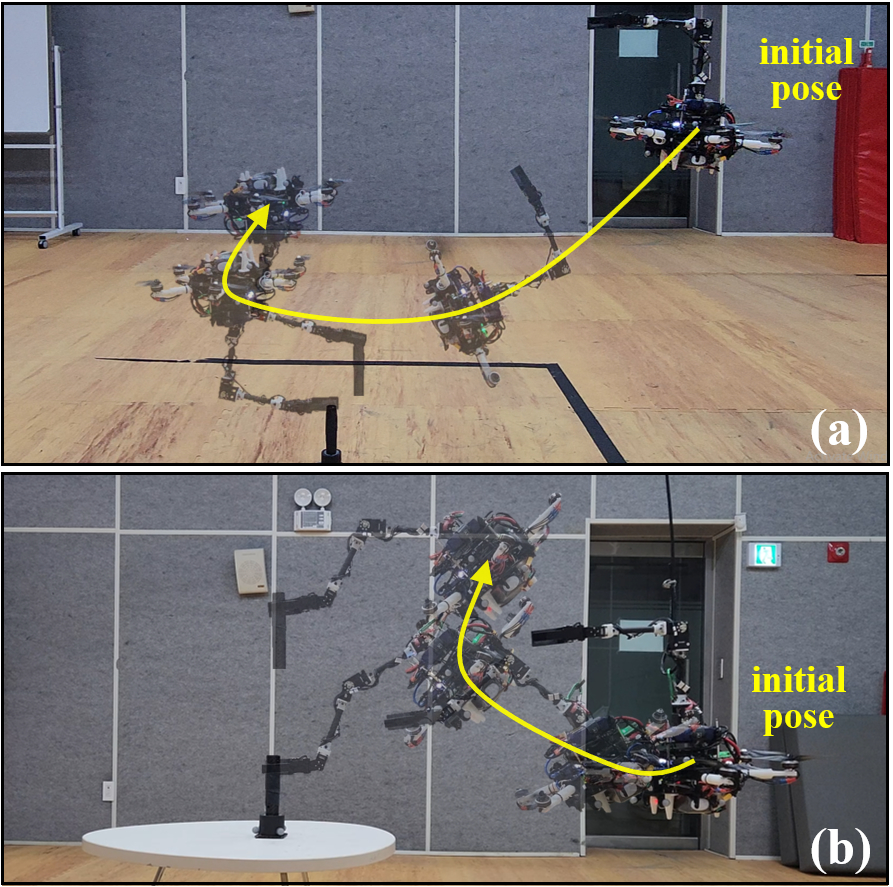}
    \caption{An omnidirectional aerial manipulator (OAM) conducting a precise manipulation task of grasping-and-pulling a bar (a) on the ground and (b) on a table. The whole-body motion is computed from the proposed cascaded motion planner, and the reference trajectory is tracked by the proposed geometric robust controller.}
    \label{fig:thumbnail}
\end{figure}

In this study, we aim to investigate a mobile manipulator that can locate the floating base in any 3D space with arbitrary position and orientation, as depicted in Fig. \ref{fig:thumbnail}. To achieve this goal, we consider an omnidirectional aerial manipulator (OAM) which is a combination of an omnidirectional multirotor \cite{allenspach2020design} and a multi-degrees-of-freedom (multi-DoF) robotic arm. Our study focuses on addressing two main issues: developing a method 1) to stably control the position and orientation of the multirotor base regardless of the arm movement or manipulation task, and 2) to enable whole-body motion planning while utilizing the omnidirectionality of the multirotor's orientation and taking into account the surrounding environment for collision avoidance and manipulation.

\subsection{Problem description} \label{sec: prob_description}
First, from a control perspective, two types of techniques are required: geometric control technique allowing the control input to be defined at arbitrary orientation on $\mathsf{SO}(3)$ and robust control technique capable of providing robustness irrespective of the arm movement or manipulation. If a local representation like Euler angles is used to express the orientation of the multirotor, it may become unstable near the $90^\circ$ pitch angle due to the singularity issue, preventing the full utilization of omnidirectionality. Additionally, if the position and orientation of the base are not properly regulated by the movement of the robotic arm or the manipulation task, it may lead to failure in sophisticated manipulation tasks. Such tracking error can result in collision with the surrounding environment, leading to a crash. Although the two issues has been tackled individually for aerial manipulators, to the best of the authors' knowledge, they have not been resolved simultaneously for OAM.

Next, from a planning perspective, it is essential to perform whole-body motion planning that considers both the position and orientation of the multirotor base, as well as the joint angles of the robotic arm. In doing so, it is necessary to: 1) consider the non-Euclidean configuration space $\mathsf{SO}(3)$ of orientation, and 2) enable online replanning.
If motion planning is conducted separately for the multirotor and robotic arm, the search space of the planning problem is restricted, potentially resulting in sub-optimality or infeasibility despite the existence of feasible solutions.
Additionally, when using local parameterization such as Euler angles to represent the orientation of the multirotor base, solution may not be obtained in certain initial conditions due to degeneracy of such local parameterization. Finally, to respond promptly to uncertainties especially during manipulation, online replanning capability is required.

\blue{For clarity, we will provide a complete formulation of these control and planning problems in section~\ref{sec:problem_formulation}.}

\subsection{Method overview and contribution}
To resolve these problems, we first propose a geometric robust integral of the tanh of the error (gRITE) controller. The proposed controller augments an integral of the tanh of the error term to a geometric nonlinear PID controller \cite{goodarzi2013geometric} which has been applied to most existing OAMs \cite{planner_Omni_OfflineOnly, zhao2023versatile, bodie2021dynamic}, and stability of the closed-loop system is formally analyzed. The added integral term is shown to be effective in suppressing the ultimate error bound, which can be made arbitrarily small with sufficiently large control gain. Next, we propose a two-step trajectory-optimization-based whole-body motion planning method. There exist two main computation bottlenecks in applying optimization-based algorithms which may hinder real-time computation and even convergence of the solution: 1) high-dimensional, non-convex search space and 2) non-Euclidean space of $\mathsf{SO}(3)$.\footnote{Simultaneously considering not only the position and orientation of the floating base but also the joint angles of the robotic arm increases the dimension of the search space, and the non-convexity comes from collision avoidance constraints.} Moreover, to represent the orientation of the base in a non-Euclidean space, nonlinear parameterizations such as quaternions or rotation matrices must be employed. 
To overcome these bottlenecks, we divide the problem into two steps where the first step solves an end-effector level subproblem, and the whole body motion considering end-effector trajectory tracking, whole-body kinematics, and physical constraints is computed in the second step.

We conduct comparative experiments for the proposed controller where the proposed one outperforms the counterpart (i.e. geometric nonlinear PID controller) in regulating both position and orientation in the presence of the robotic arm's motion. Then, the proposed control and planning framework is experimentally validated where an omnidirectional aerial manipulator (OAM) (Fig. \ref{fig:control_allocation}) performs grasping-and-pulling tasks in various environments requiring precise manipulation. Mobile manipulation using the OAM is successfully executed in five different scenarios: 1) ground-basic, 2) ground-yaw, 3) ground-pitch, 4) table-far, and 5) table-close which are illustrated in Figs. \ref{fig:ground-all}, \ref{fig:table-all}. Through the experiments, we demonstrate precise control performance regardless of disturbance, such as the movement of the robotic arm and ground effect, at arbitrary position and orientation of the base. Furthermore, the proposed planning algorithm is shown to effectively utilize whole-body motion including omnidirectionality of the floating base in conducting the manipulation task while taking into account multiple physical constraints in real-time faster than $10$ \si{Hz}.

In summary, the main contribution of this work can be summarized as follows:
\begin{itemize}
    \item We present a gRITE (geometric Robust Integral of the Tanh of the Error) controller for an omnidirectional multirotor base to enable precise mobile manipulation of the OAM. We formally prove that the proposed controller guarantees arbitrarily small ultimate bound with sufficiently large control gains.
    \item We present a two-step trajectory-optimization-based whole-body motion planning method for an omnidirectional aerial manipulator with a multi-DoF robotic arm. The proposed method is capable of online replanning in a confined space faster than $10$ \si{Hz} and exploiting the entire space of $\mathsf{SO}(3)$.
    \item  
    We conduct multiple experiments where an OAM with a multi-DoF robotic arm performs grasping-and-pulling an object either on the ground or on a table, which would not be possible for a conventional underactuated aerial manipulator with the same manipulator configuration.
    They demonstrate effectiveness and applicability of the proposed controller and planner.
    
\end{itemize}


\section{Related work}

In this section, we discuss related work on control and whole-body motion planning for the developed OAM. Comparison with existing state-of-the-art works on OAMs is summarized in Table \ref{tb:related_work_comparison}.

\begin{table*}[t!]
\caption{Comparison with state-of-the-art works on omnidirectional aerial manipulators}
\label{tb:related_work_comparison}
{
\resizebox{\linewidth}{!}{
\begin{tabular}{>{\centering\arraybackslash}m{0.17\linewidth} | 
                >{\centering\arraybackslash}m{0.13\linewidth}  
                >{\centering\arraybackslash}m{0.10\linewidth} | 
                >{\centering\arraybackslash}m{0.10\linewidth} 
                >{\centering\arraybackslash}m{0.10\linewidth} | 
                >{\centering\arraybackslash}m{0.12\linewidth} 
                >{\centering\arraybackslash}m{0.10\linewidth} 
                >{\centering\arraybackslash}m{0.09\linewidth} 
                >{\centering\arraybackslash}m{0.09\linewidth} 
                }
\toprule
\multirow{2}{*}[-0.5em]{\centering Study} & \multicolumn{2}{c|}{OAM platform} & \multicolumn{2}{c|}{Control} & \multicolumn{4}{c}{Motion planning} \\
\cline{2-9}
                       & \makecell{multirotor\\(\# of actuators)} & \makecell{manipulator\\(DoF)} & \raisebox{-0.5\height}{method} & \makecell{disturbance\\rejection} &
                       \raisebox{-0.5\height}{method} & 
                       \makecell{whole-body\\kinematics} & 
                       \makecell{collision\\avoidance} & 
                       \makecell{online\\replanning} \\
\midrule
\textbf{Ours}                    & \textbf{tiltable (12)} & \textbf{serial (4)} & \textbf{gRITE} & \cmark & \textbf{two-step TO}$^\dagger$ & \cmark & \cmark & \cmark \\
\cite{bodie2021dynamic}          & tiltable (12) & parallel (3) & FF + PID  & $\triangle$$^*$ & IK$^\dagger$ & \xmark$^\ddagger$ & \xmark & \cmark \\
\cite{planner_Omni_OfflineOnly}  & tiltable (16) & serial (3) & PID & $\triangle$ & VKC$^\dagger$ & $\triangle$$^\mathsection$ & \cmark & \xmark \\
\cite{planner_Omni_MPPI}         & tiltable (12) & fixed (0) & impedance & \xmark & MPPI$^\dagger$ & $\triangle$$^\mathparagraph$ & \xmark & \cmark \\
\cite{zhao2023versatile}         & tiltable (16) & fixed (0) & PID & $\triangle$ & \multicolumn{4}{c}{N/A (wrench planning)} \\
\cite{bodie2021active}           & tiltable (12) & fixed (0) & force \& impedance & \xmark & \multicolumn{4}{c}{N/A (force planning)} \\
\bottomrule
\end{tabular}
}
{\footnotesize
${}^*$: FF(Feed Forward) term compensates only the disturbance from the manipulator's motion and requires the acceleration measurement.

${}^\dagger$: Trajectory Optimization (TO), Inverse Kinematics (IK), Virtual Kinematic Chain (VKC), Model Predictive Path Integral Control (MPPI)

$^\ddagger$: Whole-body motion is not jointly tackled in that only the joint angles are considered.

$^\mathsection$: Singularity issue may occur due to the use of Euler angles.

$^\mathparagraph$: No joint angle is considered, but there is a possibility of extension to OAM with a multi-DoF manipulator. 
}
}
\end{table*}

\subsection{Control}

The first objective of this study is to design a geometric robust controller for an omnidirectional aerial manipulator (OAM). Accordingly, we first review robust\blue{/adaptive} or geometric robust\blue{/adaptive} control techniques applied to \blue{existing floating base robotic systems including multirotors and aerial/space manipulators}, then investigate control methods applied to an OAM. Lastly, we discuss related work on the proposed control technique, which is the geometric robust integral of the tanh of the error (gRITE).

{For the multirotor bases without considering robotic arms, various geometric robust/adaptive controllers have been proposed. \cite{goodarzi2015geometric} introduced an adaptive controller with performance analysis, while \cite{bisheban2021geometric} extended this approach by proposing a neural network-based adaptive control method that handles not only constant parametric but also state-dependent uncertainties. Further advancing this direction, \cite{o2022neural} presented a framework that integrates learning-based and adaptive control techniques to accommodate state-independent uncertainties, such as wind disturbances. However, these adaptive control approaches still exhibit performance degradation when encountering time-varying uncertainties that are not included in the training data or lack parametric modeling. Recently, \cite{wu2025L1quad} proposed an $\mathcal{L}_1$ adaptive control method that ensures the ultimate boundedness of state errors even in the presence of time-varying uncertainties. Nevertheless, achieving a sufficiently small ultimate bound with this method requires a high cutoff frequency for the accompanying low-pass filter and high gain control for state errors. As a result, in practical experiments, the performance is likely to be limited by measurement noise.
}

Next, \blue{aerial/space manipulators inherently involve dynamic coupling effect between a floating base and a rigidly attached robotic arm \cite{kim2017robust}}, and the base can become unstable if this coupling effect is not suitably addressed in a controller \cite{huber2013first}. To resolve this problem, various studies on a conventional multirotor-based aerial manipulator \cite{lee2022rise,kim2017robust,lee2021aerial,liang2023adaptive} design robust controllers where the motion of the robotic arm is treated as external disturbance. However, their use of Euler angles in attitude control renders the controllers numerically unstable near $90^\circ$ pitch angle, limiting their applicability to an OAM. Although \cite{yu2020finite} proposes a geometric robust controller using orientation error directly defined on $\mathsf{SO}(3)$, its validation is limited to simulation, and its performance may degrade when applied to a real robot due to the control law not being Lipschitz continuous. \blue{Similar technical challenges of handling uncertainties have been considered in the space robotics field where the bases are satellite or spacecraft freely floating in space \cite{zhu2019high,jia2020finite,yao2021adaptive}. However, these studies either assume planar dynamics or represent the attitude using Euler angles. \cite{xu2024predefined} tackled the control problem of a space manipulator with singularity-free attitude representation, but only ideal dynamics without uncertainties is considered.}


Studies on an OAM utilize a controller considering the non-Euclidean property of $\mathsf{SO}(3)$ to fully harness the omnidirectionality of the robot. However, as they consider only an OAM with a zero-DoF robotic arm \cite{bodie2021active,planner_Omni_MPPI,cuniato2023learning}, robustness against the relative motion of the robotic arm to the multirotor base is not tackled in their controllers. An OAM with a multi-DoF parallel robotic arm is considered in \cite{bodie2021dynamic}, and the authors design a controller with a feedforward term to compensate for the dynamic coupling effect of the robotic arm. However, as the method requires a precise dynamical model of the coupling effect and acceleration measurement, the performance may degrade when external disturbance (e.g. interaction wrench while grasping an object of an unknown mass) or excessive measurement noise exists. \cite{planner_Omni_OfflineOnly} presents a nonlinear PID controller for an OAM equipped with a multi-DoF serial robotic arm, but its performance in disturbance attenuation may be insufficient for precise manipulation tasks like grasping-and-pulling. 

This study proposes a gRITE controller, which is an extension to a RISE (robust integral of the sign of the error) controller first presented in \cite{xian2004continuous}. The RISE control achieves asymptotic stabilizability for an uncertain system using a continuous control input, and it has been adopted in various platforms \cite{kamaldin2019novel,deng2021asymptotic,shin2011autonomous}. However, due to the derivative of the control input not being continuous, input chattering may occur when applying high gains. Accordingly, in the hardware perspective where desired force/torque is tracked by actuators, the actuators’ tracking performance may deteriorate, resulting in overall performance degradation. To alleviate such problem, \cite{kidambi2021robust} replaces the sign function with the tanh function when validating its controller in simulation, but analysis for the replaced tanh function is not conducted. The original stability analysis with the sign function can no longer be applied to our case where the sign function is substituted by the tanh function because there appears a residual term in the derivative of the Lyapunov function that cannot be shown to be non-positive. \cite{xian2016new} presents formal stability analysis for a smooth counterpart of the RISE control using the tanh function, but the analysis is conducted only in $\mathbb{R}^n$ space. A geometric RISE controller is designed in \cite{gu2022agile} and is applied to a quadrotor, but its analysis is built upon a lemma that is not applicable to a case with external disturbance (Lemma 1 showing boundedness of the attitude error function), and the input chattering issue that may occur by high gains is not taken into account.

\subsection{Whole-body motion planning}

Our second objective is to present a whole-body motion planner exploiting the omnidirectionality of an OAM while abiding by various state constraints such as collision avoidance.
Considering this objective, we first review whole-body planning techniques for ground robot-based mobile manipulators. Then, we investigate planning algorithms designed for aerial manipulators including OAMs.
Lastly, we discuss related work on optimization-based motion planning where the whole configuration space of orientation, i.e. $\mathsf{SO}(3)$, is addressed.



A whole-body motion generation algorithm developed for a mobile manipulator whose base is a planar wheeled robot can be found in \cite{planar_platform1}. Planar mobile robots only demand single scalar values representing heading angles to fully describe the orientation whereas at least 3 variables are needed for the OAM's base. Thus, it is not directly applicable to the OAM. 
Quadruped robots have widely been used as bases of mobile manipulators \cite{sleiman2021unified, arcari2023bayesian, planner_WB_MPC2}.
However, since they conduct mobile manipulation while maintaining the pedals to have a stable contact with the ground, roll and pitch angles of the base orientation is restricted below $90^\circ$.
Accordingly, they rely on Euler angles to represent the base orientation. 
\blue{Another widely studied mobile manipulation platform is huamnoids. \cite{humanoid_RRT_IK, humanoid_DUAL_} leverage mobility such as standing up from a seated posture, and \cite{humanoid_loco_manipulation} incorporates bipedal locomotion for loco-manipulation. However, they either do not fully exploit bipedal walking, or when whole-body motion is considered, computation time is on the order of seconds even when using Euler angles due to the high-dimensional state space.
If the full $\mathsf{SO(3)}$ space were considered, the computational complexity would increase abruptly, making real-time replanning practically infeasible.}
Meanwhile, the OAM is capable of hovering with arbitrary base pose. Thus, to fully utilize such capability during manipulation while applying the methods for quadruped robots, one should additionally consider the entire $\mathsf{SO}(3)$ and the corresponding nonlinear kinematics.

Similarly, existing studies on aerial manipulators \cite{lee2020aerial,lee2017estimation} have limitations in that they utilize Euler angles in designing a motion planner. Although there exist studies employing the whole configuration space of $\mathsf{SO}(3)$ \cite{planner_OAM_sampling_1,welde2021dynamically}, their approaches do not tackle state constraints including obstacle and self-collision avoidance and are only validated in simulation.
As presented in the previous subsection, many studies on OAM consider a platform equipped with a zero-DoF manipulator \cite{bodie2021active,planner_Omni_MPPI,cuniato2023learning}. Among these, only \cite{planner_Omni_MPPI} tackles a problem of motion planning for OAM, but manipulator-related constraints including self-collision avoidance are not considered. 
Recent work \cite{planner_Omni_OfflineOnly} also investigates whole-body motion planning for an OAM with a multi-DoF manipulator. However, the authors represented OAM's base orientation using only three angles, leading to singularity issues, and the method can only be computed offline.



\blue{Several studies have explored trajectory generation on $\mathsf{SE(3)}$ for different robotic platforms. Quadrotors have been studied in \cite{wehbeh2022mpc, sun2022comparative}, while \cite{brescianini2018computationally} focused on fully actuated multirotors. \cite{meduri2023biconmp} studied humanoids and quadrupeds, and \cite{kalabic2017mpc} addressed motion planning for spacecraft and omnidirectional multirotors. Although these approaches enable real-time replanning, their applicability can be limited when handling collision avoidance which is typically fomulated as non-convex constraints.}



\blue{On the other hand, there exist studies that incorporate collision-free trajectory generation on $\mathsf{SE(3)}$. \cite{ding2021representation, liu2025variational} can generate a reference trajectory while ensuring collision avoidance by modeling the entire system as a single rigid body. However, this simplification yields an overly conservative behavior of systems with high-DoF robotic manipulators, as the rigid-body model must cover all possible reachable regions of the arm.
\cite{SE3_traj_Gen_with_coll_avoid_quadrupped} considers foot collision avoidance using a quaternion-based representation but neglects the base, which is critical in omnidirectional aerial manipulation.
While \cite{SE3_traj_Gen_with_coll_avoid_IJRR} proposed a platform-agnostic framework, its high computational load prevents real-time implementation.
Since aerial manipulators operate in free space, they are highly susceptible to disturbances and require rapid replanning to react to potential disturbances. 
}



{\color{black}
\section{Problem formulation} \label{sec:problem_formulation}
This section presents objectives and corresponding formulation for each control and planning problems. Also, we introduce notations used throughout the paper in Table~\ref{tab:notation}.

\begin{table}[t]
    \caption{\blue{List of notations and variables}}
    \footnotesize
    \centering
    \renewcommand{\arraystretch}{1.2} 
    \begin{tabular}{c|p{6.6cm}} 
        \toprule
        \multicolumn{1}{c|}{\textbf{Notations}} & \multicolumn{1}{c}{\textbf{Definition}} \\ 
        \midrule
        $[\bm{a};\bm{b}]$ & $[\bm{a}^\top \ \bm{b}^\top]^\top$ with two vectors $\bm{a}$ and $\bm{b}$ \\ 
        $a_i$ & $i^{th}$ element of a vector $\bm{a}$ \\ 
        $W_i$ & $i^{th}$ element of a diagonal matrix $\bm{W}$ \\ 
        $\lVert \bm{v} \rVert$ & Euclidean norm of a vector $\bm{v}$ \\ 
        $\lVert \bm{v} \rVert_{\bm{P}}^2$ & $\bm{v}^\top\bm{P}\bm{v}$ with a vector $\bm{v}$ and a positive definite matrix $\bm{P}$ \\ 
        $\bm{I}_n$ & Identity matrix in $\mathbb{R}^{n\times n}$ \\ 
        $(\cdot)^{\wedge}$ & Matrix representation of the vector cross product in $\mathbb{R}^3$ \\ 
        $(\cdot)^{\vee}$ & Inverse of the hat map \\ 
        $c*\text{, }s*$ & $\cos(*) \text{, }\sin(*)$ \\ 
        $\mathcal{A}\bigoplus \mathcal{B}$ & Minkowski sum between two sets $\mathcal{A} \text{ and } \mathcal{B}$ \\ 
        $\mathcal{S}^c$ & Complement of a set $\mathcal{S}$ \\ 
        $\Tanh(\bm{v})$ & Element-wise tanh of a vector $\bm{v}$ \\ 
        $\text{tr}(\bm{A})$ & Trace of a square matrix $\bm{A}$ \\ 
        $\lambda_m(\bm{A})$ & Minimum eigenvalue of a square matrix $\bm{A}$ \\
        \bottomrule
        \toprule
        \multicolumn{1}{c|}{\textbf{Variables}} & \multicolumn{1}{c}{\textbf{Definition}} \\ 
        \midrule
        $\bm{p}$ & Position of the multirotor base \\ 
        $\bm{R}$ & Orientation of the multirotor base \\ 
        $\bm{\omega}$ & Body angular velocity of the multirotor base \\ 
        $\bm{\theta}$ & Joint angles of the manipulator\\ 
        $^{E}\bm{p} \text{, }^{E}\bm{v}$ & End-effector position and linear velocity \\ 
        $^{E}\bm{R} \text{, }^{E}\bm{\omega}$ & End-effector orientation and angular velocity \\ 
        $\bm{d}_t \text{, } \bm{d}_r$ & External disturbance in translation and rotational dynamics \\ 
        $m \text{, } \bm{J_b}$ & True mass and MoI$^*$ of the aerial manipulator \\ 
        $\bar{m} \text{, } \bar{\bm{J}_b}$ & Nominal mass or MoI$^*$ of the aerial manipulator \\ 
        $\bm{f} \text{, }\bm{\tau}$ & Net force and torque in body frame, i.e. control inputs \\ 
        $\bm{F} \text{, } \bm{\alpha}$ & Thrusts and servo angles of rotors \\ 
        $(\cdot)_n \text{, }(\cdot)_r$ & Nominal and robust control law \\ 
        $(\cdot)_d \text{, }(\cdot)_g$ & Desired and goal value \\ 
        $\mathcal{O}_i$ & Set of all points in $\mathbb{R}^3$ constructing $i^{th}$ obstacle \\ 
        $N_o$ & Number of obstacles \\ 
        $T_f$ & Desired reaching time to the goal used in offline planner\\ 
        $T_H$ & Time Horizon used in online planner \\ 
        $\Delta t$ & Time discretization \\ 
        $g$ & Magnitude of the gravitational acceleration \\ 
        $\bm{b}_3$ & $[0,\ 0,\ 1]^\top$ \\ 
        \bottomrule
    \end{tabular}

    \vspace{1mm} 
    \begin{minipage}{\linewidth} 
        \footnotesize MoI: Moment of Inertia
    \end{minipage}

    \label{tab:notation}
\end{table}

The controller is designed with the following objectives:
\begin{itemize}
    \item Robustness to external disturbance and model uncertainty to ensure robustness against multiple sources of disturbance including the robotic arm's motion, ground effect and model uncertainty arising from an object at the end-effector.
    \item Well-defined control law in the entire state space without singularity to fully exploit the omnidirectionality of the aerial robot which can be hindered if using local coordinates such as Euler angles for representing orientation.
\end{itemize}
As for whole-body motion planning, the objectives are:
\begin{itemize}
    \item Collision-free trajectory generation to avoid both self-collision between the manipulator and the base, and collision with static obstacles in the environment.
    \item Simultaneous exploration of the joint angles and base pose over the configuration space $\mathbb{R}^3 \times \mathsf{SO}(3) \times \mathbb{R}^n$ to maximize the feasible solution set.
    \item Fast computation of trajectories to ensure real-time replanning capability for reacting to potential uncertainties.
\end{itemize}
Based on these objectives, we define the following control and planning problems under assumptions:
\begin{control_problem} \label{control_objective}
  For any given smooth reference trajectory for the omnidirectional multirotor base (i.e. position and orientation of the base), design a geometric controller capable of bounding the state tracking errors arbitrarily small under the presence of disturbances and model uncertainties.
\end{control_problem}

\begin{planning_problem} \label{planning_objective}
  For the given goal pose of the end-effector, find a whole-body motion trajectory of the OAM (i.e. pose of the floating base and joint angles of the robotic arm) that utilizes 
  the entire configuration space $\mathcal{C} = \mathbb{R}^3 \times \mathsf{SO}(3) \times \mathbb{R}^{n}$ and is free from self-collision and avoids obstacles while enforcing the end-effector to reach the goal.  
\end{planning_problem}

\begin{assumption} \label{assumption1}
    External disturbances and their time-derivatives $\bm{d}_t,\dot{\bm{d}}_t, \bm{d}_r,\dot{\bm{d}}_r$ are continuously differentiable and bounded, and $\ddot{\bm{d}}_t, \ddot{\bm{d}}_r$ are bounded.
\end{assumption}

\begin{assumption} \label{planner: AssumeObstaclesEllipsoids}
    All obstacles are static, known, and modeled as composites of multiple ellipsoids.
\end{assumption}
\begin{assumption} \label{planner: AssumeGoalRechable}
    There exists at least one curve in $\mathcal{OBS}^{\mathsf{c}} \subset \mathbb{R}^3$ starting from any points in $\mathcal{OBS}^{\mathsf{c}}$ to the goal position in $\mathcal{OBS}^{\mathsf{c}}$ where the set $\mathcal{OBS}$ is defined as 
    $\mathcal{OBS} =  \bigcup_{i=1} ^{N_o} \mathcal{O}_i$.
\end{assumption}
\noindent The first assumption is about the boundedness of the external disturbance and its time-derivatives, which is common among other papers tackling robust control against time-varying disturbances \cite{kim2017robust,lee2022rise,hua2021novel}.
The second is to allow us to formulate collision avoidance constraints with obstacles using finite dimensional parameters. 
The last assumption can be interpreted as the goal not being positioned within a fully obstructed region enclosed by obstacles. This is to ensure existence of a path from any initial position to the goal position of the end-effector.


To address the control problem, a geometric robust integral tanh of the error (gRITE) controller is proposed, and we show that the proposed method guarantees an arbitrarily small state error bound. For the planning problem, a two-step trajectory-optimization-based whole-body motion planning method is introduced, which considers the entire configuration space while maintaining real-time performance ($>10$ \si{Hz}). The flow between the algorithmic and hardware components of the proposed framework is illustrated in Fig.~\ref{fig:ExperimentsFlowDiagram-all}.

}

\begin{figure*}[t!]
    \centering
    \includegraphics[width=0.9\linewidth]{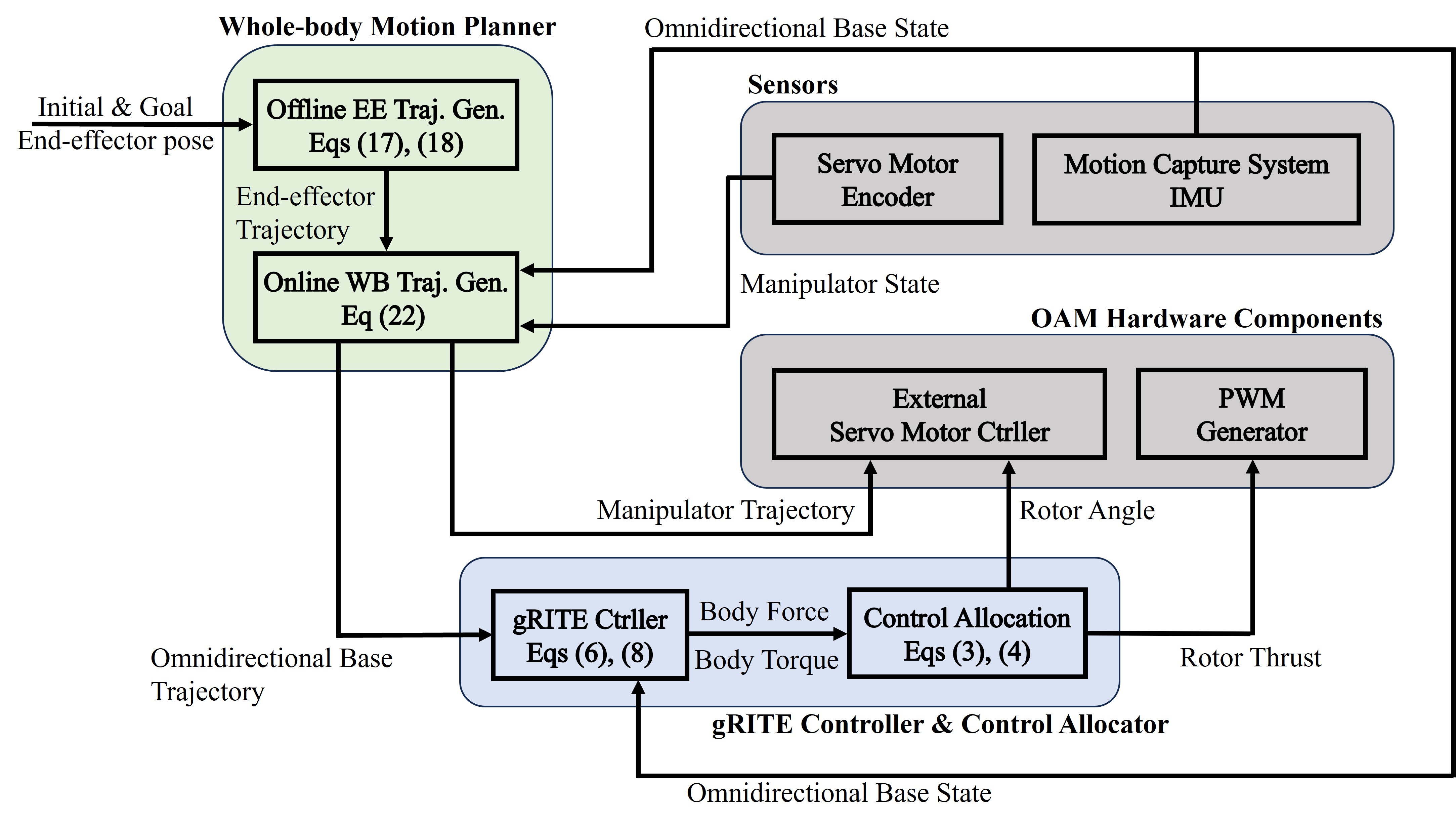}
    \caption{Overall algorithm flow for hardware experiments. EE and WB stand for End-effector and Whole-body, respectively.}
    \label{fig:ExperimentsFlowDiagram-all}
\end{figure*}
\section{Controller design}


In this section, we propose a \textit{geometric Robust Integral Tanh of the Error} (gRITE) controller which guarantees ultimate boundedness of the closed-loop system with arbitrarily small ultimate bound by choosing proper control gains.

\subsection{System dynamics} \label{subsec:system dynamics} 
We consider the following system dynamics:
\begin{subequations} \label{eq:system dynamics}
\begin{align}
    \begin{split} \label{eq:system dynamics - translation}
        m \ddot{\bm{p}} &= \bm{R}\bm{f} - mg\bm{b}_3 + \bm{d}_t
    \end{split} \\
    \begin{split} \label{eq:system dynamics - rotation}
        \bm{J}_b \dot{\bm{\omega}} &= -\bm{\omega}^{\wedge} \bm{J}_b \bm{\omega} + \bm{\tau} + \bm{d}_r
    \end{split}
\end{align}
\end{subequations}

(\ref{eq:system dynamics - translation}) models translational dynamics and (\ref{eq:system dynamics - rotation}) is for rotational dynamics.
The motion of the robotic arm induces a change in the moment of inertia of the aerial manipulator $\bm{J}_b$ and reaction force and torque to the multirotor, and these are considered as model uncertainty in $\bm{J}_b$ and external disturbance $\bm{d}_t$ and $\bm{d}_r$, respectively. If the end-effector of the robotic arm grasps an object, then the added inertia of the attached object is accounted as uncertainty in $m$ and $\bm{J}_b$ since the added inertia is generally unknown. Control inputs are force $\bm{f} \in \mathbb{R}^3$ and torque $\bm{\tau} \in \mathbb{R}^3$ represented in the multirotor body frame, indicating fully actuatedness.

\subsection{Control allocation} \label{subsec:control allocation}
\begin{figure}
    \centering
    \includegraphics[width=0.9\linewidth]{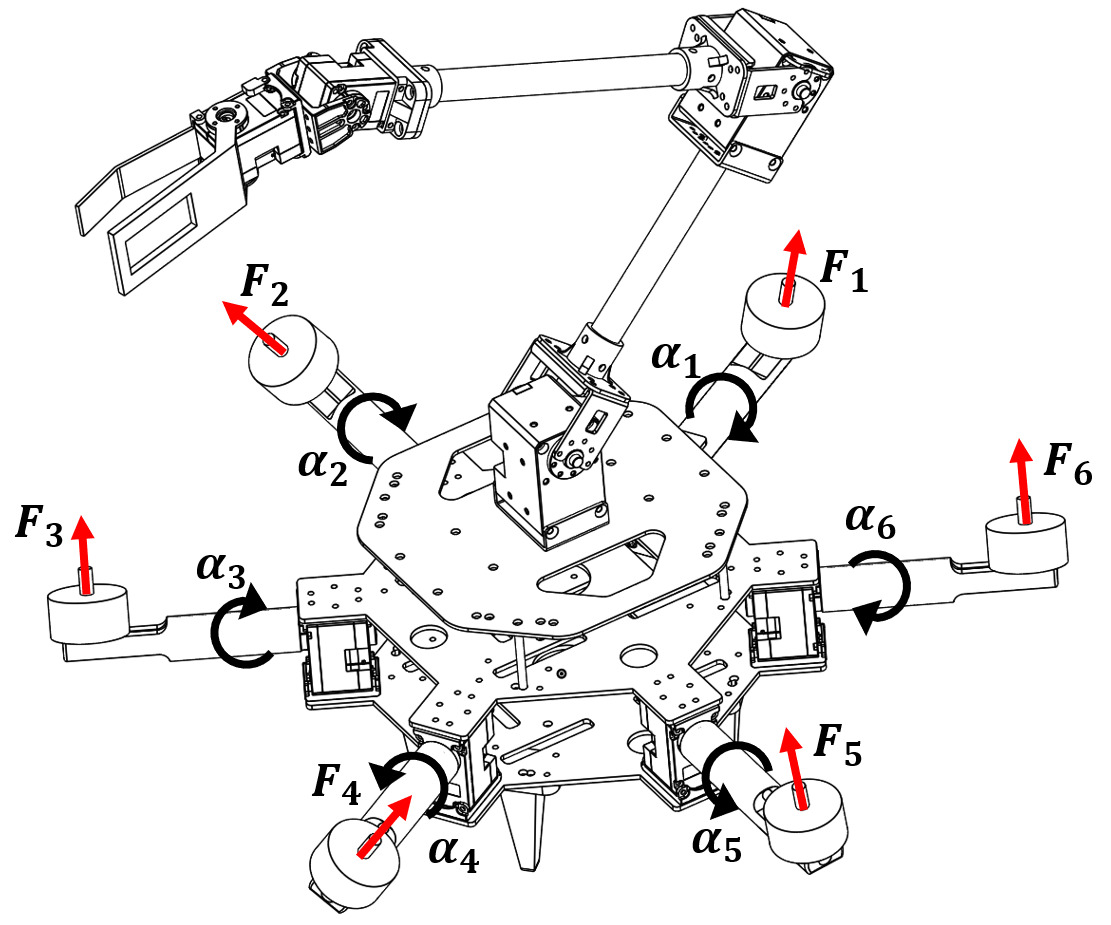}
    \caption{CAD for the omnidirectional aerial manipulator. Actuator inputs are visualized which are rotor thrust $\bm{F} = [F_1,\cdots,F_6]^\top$ and servo angle $\bm{\alpha} = [\alpha_1,\cdots,\alpha_6]^\top$.}
    \label{fig:control_allocation}
\end{figure}

The OAM considered in this paper is illustrated in Fig. \ref{fig:control_allocation}. The omnidirectional multirotor base consists of six rotors and six servomotors. As visualized in Fig. \ref{fig:control_allocation}, actuator inputs are $F_i \in \mathbb{R}_{>0}$ and $\alpha_i \in \mathbb{R}$, which are thrust of the $i^{th}$ rotor and servo angle of the $i^{th}$ servomotor, respectively. To realize a control input $[\bm{f};\bm{\tau}] \in \mathbb{R}^{6}$ in the multirotor, control allocation is required which solves a problem of finding actuator inputs $\bm{F}=[F_1,\cdots,F_6]^\top \in \mathbb{R}^6$ and $\bm{\alpha}=[\alpha_1,\cdots,\alpha_6]^\top \in \mathbb{R}^6$
for given control input $[\bm{f};\bm{\tau}]$. Using the variable transformation method in \cite{kamel2018voliro}, the relationship between the actuator inputs $\bm{F},\bm{\alpha}$ and control input $[\bm{f};\bm{\tau}]$ can be written as
\begin{equation} \label{eq: control allocation}
    \left[\begin{matrix} \bm{f} \\ \bm{\tau} \end{matrix} \right] = \bm{A} \bm{b}, \quad \bm{b} = \left[\begin{matrix} F_1 c\alpha_1 \\ F_1 s\alpha_1 \\ \vdots \\ F_6 c\alpha_6 \\ F_6 s\alpha_6     \end{matrix} \right]
\end{equation}
where $\bm{A} \in \mathbb{R}^{6 \times 12}$ is a constant matrix, whose definition can be found in Appendix \ref{appendix-A}. Such equation is derived from the geometric arrangement of actuators in the multirotor base. We compute the vector $\bm{b}$ using a weighted pseudo-inverse of $\bm{A}$ as 
\begin{equation} \label{eq: control allocation - pseudo inverse}
    \bm{b} = \bm{A}_W^\dagger \left[\begin{matrix} \bm{f} \\ \bm{\tau} \end{matrix} \right]
\end{equation}
where $\bm{A}_W^\dagger = \bm{W} \bm{A}^\top (\bm{A} \bm{W} \bm{A}^\top)^{-1}$ for a positive definite weight matrix $\bm{W} \in \mathbb{R}^{12 \times 12}_{>0}$. Then, the actuator inputs $F_i, \alpha_i$ $\forall i = 1,\cdots,6$ are calculated as 
\begin{equation}
\begin{aligned}
    F_i &= \sqrt{b_{2i-1}^2 + b_{2i}^2} \\
    \alpha_i &= \text{atan2}\left(b_{2i},b_{2i-1}\right).
\end{aligned}
\end{equation}

\subsection{The proposed control law: gRITE controller}
In this subsection, we construct control laws for $\bm{f}$ in the translational dynamics (\ref{eq:system dynamics - translation}) and $\bm{\tau}$ in the rotational dynamics (\ref{eq:system dynamics - rotation}).
\subsubsection{Translational dynamics}
We design a controller for the translational part first. We define error variables as 
\begin{equation} \label{eq:error variables - translation}
\begin{aligned}
    \bm{e}_p &= \bm{p}_d - \bm{p}, &\quad \bm{e}_{t1} &= \dot{\bm{e}}_p + \bm{\Lambda}_{t} \bm{e}_p  \\
    \bm{e}_{t2} &= \dot{\bm{e}}_{t1} + \bm{e}_{t1}, &\quad \bm{e}_t &= [\bm{e}_p;\bm{e}_{t1};\bm{e}_{t2}]
\end{aligned}
\end{equation}
where $\bm{\Lambda}_t \in \mathbb{R}^{3\times 3}_{>0}$ is a diagonal control gain.
Then, the proposed gRITE controller for the translational dynamics is formulated as follows:
\begin{subequations} \label{eq:ctrller - translation}
\begin{align}
    \begin{split} \label{eq:ctrller - translation - f}
        \bm{f} &= \bm{f}_n + \bm{f}_r     
    \end{split}\\
    \begin{split} \label{eq:ctrller - translation - fn}
        \bm{f}_n &=  \bar{m} \bm{R}^\top (g \bm{b}_3 + \bm{K}_{tp} \bm{e}_p + \bm{K}_{td} \dot{\bm{e}}_p + \ddot{\bm{p}}_d)    
    \end{split}\\
    \begin{split} \label{eq:ctrller - translation - fr}
        \bm{f}_r &= \bm{R}^\top \Big\{(\bm{K}_{ti}+\rho_{t} \bm{I}_3) (\bm{e}_{t1}(t) - \bm{e}_{t1}(0)) + \\
        &\quad \int_0^t (\bm{K}_{ti}+\rho_{t} \bm{I}_3) \bm{e}_{t1}(\tau) + \bm{\Gamma}_t \text{Tanh}(\bm{\Theta}_t \bm{e}_{t1}(\tau)) d\tau
    \Big\}    
    \end{split}
\end{align}    
\end{subequations}
$\bm{K}_{tp},\bm{K}_{td},\bm{K}_{ti},\bm{\Gamma}_{t},\bm{\Theta}_{t} \in \mathbb{R}^{3\times3}_{>0}$ and $\rho_t \in \mathbb{R}_{>0}$ are diagonal control gains. 
Briefly speaking, the nominal control law $\bm{f}_n$ exponentially stabilizes the nominal dynamics (i.e. (\ref{eq:system dynamics - translation}) with nominal parameters and no disturbance $\bm{d}_t$) while the robust control law $\bm{f}_r$ mitigates the gap between the nominal and actual dynamics. 

\subsubsection{Rotational dynamics}
Similar to the controller design for the translational dynamics, we take error variables first in designing a controller for the rotational counterpart as follows:
\begin{equation} \label{eq:error variables - rotation}
\begin{gathered}
    \bm{e}_R = \frac{1}{2}(\bm{R}^\top \bm{R}_d - \bm{R}_d^\top \bm{R})^{\vee}, \quad \bm{e}_\omega = \bm{R}^\top \bm{R}_d \bm{\omega}_d - \bm{\omega} \\
    \bm{e}_{r1} = \bm{e}_\omega + \bm{\Lambda}_r \bm{e}_R , \quad \bm{e}_{r2} = \dot{\bm{e}}_{r1} + \bm{e}_{r1}, \quad \bm{e}_r = [\bm{e}_R;\bm{e}_{r1};\bm{e}_{r2}]
\end{gathered}
\end{equation}
The error variables for the orientation $\bm{e}_R$ and the angular velocity $\bm{e}_\omega$ are defined following \cite{lee2010geometric} where $\bm{R}_d \in \mathsf{SO}(3)$ and $\bm{\omega}_d \in \mathbb{R}^3$ are the desired orientation and body angular velocity, respectively. $\bm{\Lambda}_r \in \mathbb{R}^{3\times 3}_{>0}$ is a diagonal control gain. Then, the proposed gRITE control law for the rotational dynamics is as follows:
\begin{subequations} \label{eq:ctrller - rotation}
\begin{align}
    \begin{split} \label{eq:ctrller - rotation - tau}
        \bm{\tau} &= \bm{\tau}_n + \bm{\tau}_r    
    \end{split} \\
    \begin{split} \label{eq:ctrller - rotation - taun}
        \bm{\tau}_n &= \bm{\omega}^\wedge \bar{\bm{J}}_b \bm{\omega} - \bar{\bm{J}}_b (\bm{\omega}^\wedge \bm{R}^\top \bm{R}_d \bm{\omega}_d - \bm{R}^\top \bm{R}_d \dot{\bm{\omega}}_d) +\\
        &\quad \bar{\bm{J}}_b \bm{K}_{rp} \bm{e}_R + \bar{\bm{J}}_b \bm{K}_{rd} \bm{e}_{\omega}
    \end{split} \\
    \begin{split} \label{eq:ctrller - rotation - taur}
        \bm{\tau}_r &= (\bm{K}_{ri}+\rho_r \bm{I}_3) (\bm{e}_{r1}(t) - \bm{e}_{r1}(0)) + \\
        &\quad \int_0^t (\bm{K}_{ri}+\rho_r \bm{I}_3) \bm{e}_{r1}(\tau) + \bm{\Gamma}_r \text{Tanh}(\bm{\Theta}_r \bm{e}_{r1}(\tau)) d\tau    
    \end{split}
\end{align}    
\end{subequations}
$\bm{K}_{rp},\bm{K}_{rd},\bm{K}_{ri},\bm{\Gamma}_{r},\bm{\Theta}_{r} \in \mathbb{R}^{3\times3}_{>0}$ and $\rho_r \in \mathbb{R}_{>0}$ are diagonal control gains, and the nominal control law $\bm{\tau}_n$ exponentially stabilizes the nominal error dynamics where the nominal moment of inertia $\bar{\bm{J}}_b$ is considered and no disturbance $\bm{d}_r$ exists \cite{lee2010geometric}.

\section{Stability analysis} \label{sec: stability analysis}
 
Using the derived control laws for the translational and rotational dynamics (\ref{eq:ctrller - translation}), (\ref{eq:ctrller - rotation}), we analyze stability of each closed-loop system. Note that the following analysis is motivated by \cite{xian2016new}. 


\subsection{Translational dynamics}
From (\ref{eq:system dynamics - translation}), (\ref{eq:ctrller - translation}), dynamics of $\bm{e}_{t2}$ can be written as
\begin{equation} \label{eq: et2 dynamics}
    m \dot{\bm{e}}_{t2} = \bm{N}_t - (\bm{K}_{ti} + \rho_t \bm{I}_3) \bm{e}_{t2} - \bm{\Gamma}_t \Tanh(\bm{\Theta}_t \bm{e}_{t1}) - \bm{e}_{t1}
\end{equation}
where 
\begin{equation*}
\begin{aligned}
    \bm{N}_t = {}&(m - \bar{m}) \dddot{\bm{p}}_d - \dot{\bm{d}}_t -\bar{m} (\bm{K}_{tp} \dot{\bm{e}}_p + \bm{K}_{td} \ddot{\bm{e}}_p) + \\
    &m(\bm{\Lambda}_t \ddot{\bm{e}}_p + \dot{\bm{e}}_{t1}) + \bm{e}_{t1} \\
    \bm{N}_{td} = {}&(m - \bar{m}) \dddot{\bm{p}}_d - \dot{\bm{d}}_t.
\end{aligned}   
\end{equation*}
Define $\tilde{\bm{N}}_t = \bm{N}_t - \bm{N}_{td} = -\bar{m} (\bm{K}_{tp} \dot{\bm{e}}_p + \bm{K}_{td} \ddot{\bm{e}}_p) + m(\bm{\Lambda}_t \ddot{\bm{e}}_p + \dot{\bm{e}}_{t1}) + \bm{e}_{t1}$, then there exist a constant $\mu_t > 0$ satsifying
\begin{equation} \label{eq: Ntilde t}
    \lVert \tilde{\bm{N}}_t \rVert \leq \mu_t \lVert \bm{e}_t \rVert.
\end{equation}

Now, we define a Lyapunov candidate function $V_t$ as
\begin{equation} \label{eq: lyapunov Vt}
    V_t = \cfrac{1}{2} \bm{e}_p^\top \bm{e}_p + \cfrac{1}{2} \bm{e}^\top_{t1}\bm{e}_{t1} + \cfrac{1}{2} m \bm{e}^\top_{t2}\bm{e}_{t2} + Q_t(t) 
\end{equation}
where 
\begin{equation*}
    Q_t = \sum^n_{i=1} \cfrac{\Gamma_{t,i}}{\Theta_{t,i}} \text{ln}(\text{cosh}(\Theta_{t,i} e_{t1,i})) - e_{t1,i} N_{td,i} + \cfrac{\Gamma_{t,i}}{\Theta_{t,i}} \text{ln} 2.
\end{equation*}
For the ease of analysis, here we also derive the time-derivative of $V_t$ as
\begin{equation*}
\begin{gathered}
    \dot{V}_t = -\bm{e}_p^\top \bm{\Lambda}_t \bm{e}_p - \lVert \bm{e}_{t1} \rVert^2 - \bm{e}_{t2}^\top (\bm{K}_{ti} + \rho_t \bm{I}_3 ) \bm{e}_{t2} + \\
    \bm{e}_p^\top \bm{e}_{t1} + \bm{e}_{t2}^\top \tilde{\bm{N}}_t + O_t(t)
\end{gathered}
\end{equation*}
where $O_t = \bm{e}_{t1}^\top(\bm{N}_{td} - \dot{\bm{N}}_{td}) - \bm{e}_{t1}^\top \bm{\Gamma}_t \Tanh(\bm{\Theta}_t \bm{e}_{t1}).$
\begin{lemma}
    Take $\Gamma_{t,i} \geq \lVert N_{td,i} \rVert_\infty + \lVert \dot{N}_{td,i} \rVert_\infty$ $\forall i$. Then, $Q_t$ satisfies the following inequalities:
    \begin{equation*}
    \begin{aligned}
        \sum^n_{i=1}(\Gamma_{t,i}-\lVert N_{td,i} \rVert_\infty) \lvert e_{t1,i} \rvert \leq Q_t(t) \leq \\ 
        \sum^n_{i=1}(\Gamma_{t,i} + \lVert N_{td,i} \rVert_\infty) \lvert e_{t1,i} \rvert + \cfrac{\Gamma_{t,i}}{\Theta_{t,i}} \textup{ln} 2.
    \end{aligned}
    \end{equation*}
\end{lemma}
\begin{proof}
    Finite $\Gamma_{t,i}$ exists by the Assumption \ref{assumption1}, and the inequalities can be derived by using the fact that $\lvert x \rvert \leq \text{ln}(\text{cosh}(x)) + \text{ln}2 = \text{ln}(e^{x} + e^{-x})$ and $\text{ln}(\text{cosh}(x)) \leq \lvert x \rvert$ for any $x \in \mathbb{R}$.
\end{proof}
\begin{lemma}
    Take $\Gamma_{t,i} \geq \lVert N_{td,i} \rVert_\infty + \lVert \dot{N}_{td,i} \rVert_\infty$ $\forall i$. Then, with $c=0.2785$, the upper bound of $O_t$ can be obtained as
    \begin{equation*}
        O_t \leq \sum^n_{i=1} \Gamma_{t,i} \lvert e_{t1,i} \rvert (1 - \textup{tanh}(\Theta_{t,i} \lvert e_{t1,i} \rvert)) \leq \sum^n_{i=1}\cfrac{\Gamma_{t,i}}{\Theta_{t,i}} c.
    \end{equation*}
\end{lemma}
\begin{proof}
    Finite $\Gamma_{t,i}$ exists by Assumption \ref{assumption1}, and the upper bound can be obtained by using the fact that $\max_{x\in \mathbb{R}}\{ \lvert x \rvert - x \text{tanh}(x) \} \leq c$ \cite{polycarpou1996robust,jia2019robust} and the control gain condition on $\Gamma_{t,i}$.
\end{proof}
\begin{lemma} \label{lemma 3}
    Assume that the following scalar differential equation holds for any sufficiently smooth $\alpha(\cdot) \in \mathcal{K}_\infty$ and constant scalar $c_s$:
    \begin{equation*}
        \dot{s} \leq -\alpha(s) + \alpha(c_s).
    \end{equation*}
    Then, there exist $\beta(\cdot,\cdot) \in \mathcal{KL}$ such that the following holds:
    \begin{equation*}
        s(t) \leq \beta(s(0) - c_s,t) + c_s \quad \forall t \geq 0.
    \end{equation*}
\end{lemma}
\begin{proof}
    See the proof in Appendix \ref{appendix: proof of lemma 3}.
\end{proof}
\begin{theorem} \label{theorem 1}
    For control gains satisfying 
    \begin{equation*}
    \begin{aligned}
        \Gamma_{t,i} &\geq \lVert N_{td,i} \rVert_\infty + \lVert \dot{N}_{td,i} \rVert_\infty \quad \forall i \\   
        \lambda_m(\bm{\Lambda}_t) &\geq 0.5 \\
        \eta^*_t &\geq \mu_t^2/(2\lambda_m(\bm{K}_{ti}))
    \end{aligned}
    \end{equation*}
    where $\eta^*_t = \min \{\lambda_{m}(\bm{\Lambda}_t) - \frac{1}{2}, \frac{1}{2}, \rho_t  \}$, the closed-loop system of the translational dynamics consisting of (\ref{eq:error variables - translation}) and (\ref{eq: et2 dynamics}) is ultimately bounded, and the ultimate bound can be made arbitrarily small.
\end{theorem}
\begin{proof}
    See the proof in Appendix \ref{appendix: proof of theorem 1}.
\end{proof}
Briefly speaking, to satisfy the control gain conditions, one should take large enough $\bm{K}_{ti}$ and $\bm{\Gamma}_t$. Furthermore, the steady-state error bound can be arbitrarily shrunk by taking large enough $\bm{\Theta}_t$, and this corresponds to the fact that no steady-state error exists if the sign function is used instead of tanh \cite{xian2004continuous}. However, as will be \blue{discussed at the end of this section, using tanh instead of the sign function has advantages in attenuating input chattering.}

\subsection{Rotational dynamics}
From (\ref{eq:system dynamics - rotation}) and (\ref{eq:ctrller - rotation}), dynamics of $\bm{e}_{r2}$ can be obtained as 
\begin{equation}
\begin{aligned}
    \bm{J}_b \dot{\bm{e}}_{r2} &= -(\bm{K}_{ri} + \rho_r \bm{I}_3) \bm{e}_{r2} - \bm{\Gamma}_r \Tanh(\bm{\Theta}_r \bm{e}_{r1}) - \frac{1}{2} \dot{\bm{J}}_b \bm{e}_{r2} \\
    &\quad - \bm{e}_{r1} + \tilde{\bm{N}}_r + \bm{N}_{rd}
\end{aligned}
\end{equation}
where
\begin{equation*}
\begin{aligned}
    \bm{N}_{rd} &= (\dot{\bm{E}} - \dot{\bar{\bm{E}}}) - \dot{\bm{d}}_r, \\
    \tilde{\bm{N}}_r &= -\dot{\bar{\bm{J}}}_b (\bm{K}_{rp} \bm{e}_R + \bm{K}_{rd} \bm{e}_\omega) - \bar{\bm{J}}_b (\bm{K}_{rp} \dot{\bm{e}}_R + \bm{K}_{rd} \dot{\bm{e}}_\omega) \\
    &\quad + \bm{e}_{r1} + \frac{1}{2}\dot{\bm{J}}_b \bm{e}_{r2} + (\bm{J}_b - \dot{\bm{J}}_b) \dot{\bm{e}}_{r1} + \dot{\bm{J}}_b \bm{\Lambda}_r \bm{C} \bm{e}_\omega + \\
    &\quad \bm{J}_b \bm{\Lambda}_r (\dot{\bm{C}} \bm{e}_\omega + \bm{C}\dot{\bm{e}}_\omega), \\
    \bm{E} &= \bm{J}_b (\bm{R}^\top \bm{R}_d \dot{\bm{\omega}}_d - \bm{\omega}^{\wedge} \bm{R}^\top \bm{R}_d \bm{\omega}_d), \\
    \bm{C} &= \frac{1}{2} [\text{tr}(\bm{R}^\top \bm{R}_d) \bm{I}_3 - \bm{R}^\top \bm{R}_d].
\end{aligned}
\end{equation*}
Here, we use the fact that $\dot{\bm{e}}_R = \bm{C} \bm{e}_{\omega}$ where $\lVert \bm{C} \rVert_2 \leq 1$ for any $\bm{R}^\top \bm{R}_d \in \mathsf{SO}(3)$, and $\bar{\bm{E}} = \bm{E} \rvert_{\bm{J}_b = \bar{\bm{J}}_b}$. From Remark 3 in \cite{xian2004continuous}, there exists $\mu_r(\cdot) \in \mathcal{K}_\infty$ such that $\lVert \tilde{\bm{N}}_r \rVert \leq \mu_r(\lVert \bm{e}_r \rVert) \lVert \bm{e}_r \rVert$ holds.

Now, we define a Lyapunov candidate function $V_r$ as
\begin{equation} \label{eq: lyapunov Vr}
    V_r = \frac{1}{2} \bm{e}^\top_{r1} \bm{e}_{r1} + \frac{1}{2} \bm{e}^\top_{r2}\bm{J}_b \bm{e}_{r2} + Q_r + \Psi
\end{equation}
where
\begin{equation*}
\begin{aligned}
    Q_r &= \sum^n_{i=1} \cfrac{\Gamma_{r,i}}{\Theta_{r,i}} \text{ln}(\text{cosh}(\Theta_{r,i} e_{r1,i})) - e_{r1,i} N_{rd,i} + \cfrac{\Gamma_{r,i}}{\Theta_{r,i}} \text{ln} 2 \\
    \Psi &= \frac{1}{2}\text{tr}[\bm{I}_3 - \bm{R}^\top \bm{R}_d].
\end{aligned}    
\end{equation*}
Note that to capture non-Euclidean property of $\mathsf{SO}(3)$, $\Psi$ is introduced in (\ref{eq: lyapunov Vr}) instead of {$\bm{e}_p^\top \bm{e}_p$} in (\ref{eq: lyapunov Vt}).
\begin{lemma} \label{lemma 4}
    Assume that control gains satisfy the following: 
    \begin{equation*}
    \begin{aligned}
        \lambda_m(\bm{K}_{ri}) &\geq \frac{1}{2\eta_r^*} \mu_r^2\left(\sqrt{\frac{1}{\eta_r} \beta_r(\lvert V_r(0) - \Xi_r \rvert,t) + \frac{\Xi_r}{\eta_r}} \right) \\
        \lambda_m(\bm{\Lambda}_r) &\geq 0.5 \\
        \Gamma_{r,i} &\geq \lVert N_{rd,i} \rVert_\infty + \lVert \dot{N}_{rd,i} \rVert_\infty \quad \forall i
    \end{aligned}
    \end{equation*}
    where $\eta^*_r = \min \{\lambda_m(\bm{\Lambda}_r)-\frac{1}{2},\frac{1}{2}, \rho_r \}$.
    If $\Psi(t) \leq \psi$ holds for some $0< \psi < 2$ and $t \in [t_1, t_2]$ for any $0 \leq t_1 < t_2$, then the following holds for all $t \in [t_1,t_2]$:
    \begin{equation} \label{eq: Vr omega bound}
        \dot{V}_r \leq -\Omega_r(V_r) + \Omega_r(\Xi_r). 
    \end{equation}
    Here, $\beta_r(\cdot,\cdot) \in \mathcal{KL}$, and $\Xi_r = \chi(\sigma_r)$ for some $\chi(\cdot) \in \mathcal{K}_\infty$ and $\sigma_r = \sum{ \frac{\Gamma_{r,i}}{\Theta_{r,i}} \textnormal{ln}2}$ whose detailed definitions are in Appendix \ref{appendix: proof of lemma 4}.
\end{lemma}
\begin{proof}
    See the proof in Appendix \ref{appendix: proof of lemma 4}.
\end{proof}

\begin{theorem} \label{theorem 2}
    Assume that control gains $\bm{\Theta}_r, \bm{\Gamma}_r$ are selected so that $\Xi_r < \psi$ holds for some $\psi \in (0,2)$. We also assume that control gain conditions in Lemma 2 hold, and the initial condition satisfies
    \begin{equation} \label{eq: initial condition - rotation}
        V_r(0) \leq \psi - \epsilon_\psi
    \end{equation}
    for arbitrarily small positive scalar $\epsilon_\psi$.
    Then, the closed-loop system of the rotational dynamics consisting of (\ref{eq:system dynamics - rotation}) and (\ref{eq:ctrller - rotation}) is ultimately bounded, and the ultimate bound can be made arbitrarily small.
\end{theorem}
\begin{proof}
    See the proof in Appendix \ref{appendix: proof of theorem 2}.
\end{proof}

Although the sign function can provide additional merit of asymptotic stability \cite{gu2022agile,xian2004continuous} if used instead of the tanh function in the control law (\ref{eq:ctrller - translation}), (\ref{eq:ctrller - rotation}), it may result in input chattering. Considering the omnidirectional aerial robot where some actuators (i.e. servomotors) show unknown, non-negligible time delay in tracking the input command, such input chattering can deteriorate the tracking performance. The tanh function in the proposed control law relieves this problem by providing sufficient smoothness in the control input while endowing the necessary disturbance attenuation property with moderately high control gains of $\bm{\Theta}_t, \bm{\Theta}_r$.

As done in \cite{xian2016new}, gRITE control only with the robust control law $\bm{f}_r$ and $\bm{\tau}_r$ can provide the same property of boundedness with arbitrary small bound in theory. However, since the robust control law does not exploit the structure of the system dynamics, feedforward terms such as gravity $\bar{m} g \bm{R}^\top \bm{b}_3$ in (\ref{eq:ctrller - translation - fn}) and centrifugal force $\bm{\omega}^{\wedge} \bar{\bm{J}}_b \bm{\omega}$ in (\ref{eq:ctrller - rotation - taun}) cannot be utilized, and this may degrade initial transient performance. Furthermore, since proportional and derivative gains for position/orientation errors $\bm{e}_p$, $\bm{e}_R$ cannot be tuned independently to the integral gains, the gain tuning process can become demanding. The nominal control law mitigates the two problems because it is designed to involve feedforward terms and has proportional, derivative gains independent to the integral gains.

\section{Whole-body motion planning}

This section overviews our whole-body trajectory generation algorithm designed for an OAM.
We utilize both offline and online planning to enhance convergence to the predefined end-effector goal pose and to enable reactive real-time replanning with respect to potential disturbance.
This allows the OAM to reach the predefined end-effector goal pose and refine the entire state trajectory in real time.
In the initial stage, our algorithm focuses on the end-effector trajectory.
The end-effector pose is modeled as a particle in a 3D Euclidean space with orientation.
As a result, the globally optimal, collision-free end-effector pose trajectory can be obtained within a few seconds even under a setting where the problem is highly nonlinear due to the rotational motion, and the trajectory horizon is longer than $10$ \si{s}.
Then, our algorithm proceeds to local online whole-body motion planning.
Here, the primary consideration is to consider the whole-body motion while tracking the previously determined optimal end-effector pose trajectory.
Before presenting algorithmic details, we rewrite the Assumption \ref{planner: AssumeObstaclesEllipsoids} mathematically.

Only for notational simplicity, we describe an obstacle with a single ellipsoid from the following. That is,
\begin{equation*}
    \begin{gathered}
        \mathcal{O}_i = 
        \mathcal{E}(\bar{\bm{p}}_i,\bar{\bm{Q}}_i) \subset \mathbb{R}^3 \\
        \mathcal{E}(\bar{\bm{p}}_i,\bar{\bm{Q}}_i) \coloneqq \{ \bm{p} \in \mathbb{R}^3| (\bm{p}-\bar{\bm{p}}_i)^\top \bar{\bm{Q}}_i^{-1} (\bm{p}-\bar{\bm{p}}_i)\leq1\}    
    \end{gathered}
    \end{equation*}
\noindent where $\bar{\bm{p}}_i\in \mathbb{R}^3$ and $\bar{\bm{Q}}_i\in \mathbb{R}^{3\times3}_{>0}$ denote the center and the shape matrix of the $i^{th}$ ellipsoid, respectively. This assumption allows us to formulate collision avoidance constraints with obstacles using finite dimensional 
parameters of $\bar{\bm{p}}_i$ and $\bar{\bm{Q}}_i$. 
When it is difficult to represent a particular obstacle as a single ellipsoid, we can construct a composite of multiple ellipsoids that fully encompass the obstacle.

Now, we formulate the following Optimal Control Problems (OCPs) for our cascaded whole-body motion planner.

\subsection{Offline end-effector trajectory generation} \label{planner:Offlinesection}
From Assumption \ref{planner: AssumeObstaclesEllipsoids}, it becomes apparent that the end-effector trajectory avoids collisions with obstacles if and only if the following inequalities are satisfied:
\begin{gather*}
    h_i(^E \bm{p}_d(k)) > 0 \quad \forall i, k \\
    h_i(^E \bm{p}_d) := (^E \bm{p}_d-\bar{\bm{p}}_i)^\top \bar{\bm{Q}}_i^{-1} (^E \bm{p}_d-\bar{\bm{p}}_i) - 1 
\end{gather*}
where $k$ denotes the discretized time index throughout this section. However, directly incorporating these constraints into discrete-time OCPs may lead to a numerical issue of generating an optimal trajectory that penetrates very thin ellipsoids within a single time interval. In such cases, the system may collide with ellipsoids although the optimized trajectory is feasible. Therefore, instead, we formulate the following collision avoidance constraints, which impose constraints on the linear velocity.
\begin{gather} \label{planner: EECollisionAvoidance}
    \tilde{h}_i(^E \bm{p}_d(k), ^{E} \bm{v}_d(k)) > 0 \quad \forall i,k \\ \nonumber
    \tilde{h}_i(^E \bm{p}_d, ^{E} \bm{v}_d) := \frac{\partial h_{i}}{\partial ^E \bm{p}_d}(^E \bm{p}_d){^{E} \bm{v}_d} + \gamma_i ( h_{i}(^{E}\bm{p}_d) )
\end{gather}
where $\gamma_i(\cdot) \in \mathcal{K}_\infty$ can be chosen arbitrarily. We designed $\gamma_i$ as linear functions for simplicity. Provided that the inequalities (\ref{planner: EECollisionAvoidance}) and $h_{i}(^{E}\bm{p}_d(0))>0$ hold, the forward invariance of $h_{i} > 0$ is guaranteed by the comparison lemma \cite[Lemma 3.4]{khalil2002nonlinear}. 
\blue{A similar technique is utilized in the framework of control barrier function \cite{CBF_TAC, CBF_review}, where the decay rate of a control barrier function is bounded so that a safe and smooth motion is obtained.}

The modified collision avoidance constraints (\ref{planner: EECollisionAvoidance}) rely only on the optimization variables in the translational motion.
Thus, we can construct the following decoupled jerk minimization problem.
\begin{equation} \label{planner: EEOCPTranslation}
\begin{aligned}
\min_{^E \bm{x}_p, ^E \ddot{\bm{v}}_d} & \sum_{k=0}^{N_T-1} { \lVert ^E \ddot{\bm{v}}_d(k) \rVert_{R_v}^2} \\
\textrm{s.t.} \quad & ^E\bm{x}_p(0) = [^{E}\bm{p}(0);\bm{0};\bm{0}] \\
   & ^E\bm{x}_p(N_T) = [{}^{E}\bm{p}_g;\bm{0};\bm{0}] \\
     \forall_k \quad & ^E\bm{x}_p(k+1)= {}^E\bm{f}_p(^E\bm{x}_p(k), ^E \ddot{\bm{v}}_d(k)) \\
   \forall_{i,k} \quad &\tilde{h}_i(^E \bm{p}_d(k), ^{E} \bm{v}_d(k)) > 0  
\end{aligned}
\end{equation}

\begin{equation} \label{planner: EEOCPRotation}
\begin{aligned}
\min_{^E \bm{x}_R, ^E \ddot{\bm{\omega}}_d} & \sum_{k=0}^{N_T-1} { \lVert ^E \ddot{\bm{\omega}}_d(k) \rVert_{R_\omega}^2} \\
\textrm{s.t.} \quad & ^E\bm{x}_R(0) = ( {}^{E}\bm{R}(0), \bm{0}, \bm{0} ) \\
   & ^E\bm{x}_R(N_T) = ({}^{E}\bm{R}_g, \bm{0}, \bm{0}) \\
   \forall_k \quad &^E\bm{x}_R(k+1)= {}^E\bm{f}_R(^E\bm{x}_R(k), ^E \ddot{\bm{\omega}}_d(k))\\
   \end{aligned}
\end{equation} 
Here, $^E\bm{x}_p:=[{}^E\bm{p}_d;{}^E\bm{v}_d;{}^E\dot{\bm{v}}_d]\in\mathbb{R}^9$, $^E\bm{x}_R := ({}^E\bm{R}_d,{}^E\bm{\omega}_d,{}^E\dot{\bm{\omega}}_d)\in\mathsf{SO}(3)\times\mathbb{R}^6$, and $N_T=T_f/\Delta t$.
$^E\bm{f}_p$ and $^E\bm{f}_R$ are time-discretized linear and nonlinear kinematics of the translational and rotational motion. The detailed definitions of $^E\bm{f}_p$ and $^E\bm{f}_R$ can be found in Appendix \ref{appendix: def of offline kinematics}.   
$\bm{R}_v$ and $\bm{R}_\omega$ are positive definite weight matrices. $^E\bm{p}(0)$ and $^E\bm{R}(0)$ are obtained from the forward kinematics with the measured values of $\bm{p}(0)$, $\bm{R}(0)$, and $\bm{\theta}(0)$ when the planner algorithm initiates.
It is worth noting that, if the objective is only related to the goal position, (\ref{planner: EEOCPRotation}) can be omitted thanks to the fact that (\ref{planner: EEOCPTranslation}) and (\ref{planner: EEOCPRotation}) are decoupled.

\subsection{Online whole-body motion planning}
Unlike the offline end-effector trajectory generation, whole-body motion planning necessitates the consideration of collision avoidance among rigid bodies, identification of the optimal configuration among those achieving the same end-effector pose, and online replanning to react to uncertainties and refine the end-effector trajectory if unavoidable.
This subsection provides details of how each of these considerations is incorporated into the OCP.
\subsubsection{Collision avoidance constraint}
Let us define the $i^{th}$ ellipsoid $\mathcal{A}_i(t)$ comprising the aerial manipulator at time $t$ as 
\begin{equation*}
\begin{gathered}
    \mathcal{A}_i(t) := \mathcal{E}({}^A\bar{\bm{p}}_i(t),{}^A\bar{\bm{Q}}_i(t)) \\ 
    {}^A\bar{\bm{Q}}_i(t) := {}^A\bm{R}_i(t) {}^A\bar{\bm{Q}}_i(0) {}^A\bm{R}_i(t)^\top
\end{gathered}
\end{equation*}
In the above, ${}^A\bar{\bm{p}}_i$ and ${}^A\bm{R}_i$ represents the position and orientation of the $i^{th}$ link obtained through the forward kinematics, where the multirotor base corresponds to $i=0$. 
${}^A\bar{\bm{Q}}_i(0)$ is the shape matrix calculated at ${}^A\bm{R}_i=\bm{I}_3$.

When obstacles and robot components are all modeled with ellipsoids, collision avoidance can be ensured if there is no intersection between each pair of robot and obstacle ellipsoids. Let us consider a set generated by the Minkowski sum of two sets of ellipsoids as $\bar{\mathcal{B}} = \mathcal{B}_1 \bigoplus (-\mathcal{B}_2)$. The condition for the absence of intersection between the two ellipsoids can be expressed as $0 \notin \bar{\mathcal{B}}$. However, in general, $\bar{\mathcal{B}}$ is not an ellipsoid. To resolve this, similar to the work done in \cite{Son_ellipsoidal_minkowski} where the most compact ellipsoid is used that encompasses $\bar{\mathcal{B}}$ through trace minimization, we can guarantee the absence of intersection between the given two ellipsoids if the following condition is satisfied \cite{Seo_ellipsoidal_minkowski_iros}.
\begin{equation} \label{planner: WBCollisionAvoidance}
\begin{gathered}
    \hat{h}(\mathcal{B}_1, \mathcal{B}_2) > 0 \\
    \hat{h}(\mathcal{B}_1, \mathcal{B}_2):= (\bar{\bm{p}}_{\mathcal{B}_1}-\bar{\bm{p}}_{\mathcal{B}_2})^\top \bar{\bm{Q}}^{-1}(\text{$\tiny{\mathcal{B}_1, \mathcal{B}_2}$})(\bar{\bm{p}}_{\mathcal{B}_1}-\bar{\bm{p}}_{\mathcal{B}_2}) - 1 \\
    \bar{\bm{Q}}(\mathcal{B}_1, \mathcal{B}_2) := \sum_{i=1}^2\sum_{j=1}^2 {\frac{\bar{\bm{Q}}_{\mathcal{B}_i}}{\sqrt{\text{tr}(\bar{\bm{Q}}_{\mathcal{B}_i})}}\sqrt{\text{tr}(\bar{\bm{Q}}_{\mathcal{B}_j})}}
\end{gathered}
\end{equation}
Hence, the collision avoidance is guaranteed if (\ref{planner: WBCollisionAvoidance}) holds for all pairs of ellipsoids of robots and obstacles for all $t \geq 0$. 

\subsubsection{Cost function}
In comparison with ground robot-based mobile manipulators, aerial manipulators are more susceptible to disturbance caused by interaction forces between the end-effector and the target object during manipulation. To mitigate this, it is imperative to secure the manipulability of the robot arm. Consequently, we incorporate the following manipulability index into our cost function:
\begin{equation} \label{planner: WBManipulabilityIndex}
    \phi_m(\bm{\theta}) := \mu_v \det( \bm{J}_{v}(\bm{\theta}) \bm{J}_{v}^\top(\bm{\theta})) + \mu_\omega \det( \bm{J}_{\omega}(\bm{\theta}) \bm{J}^\top_{\omega}(\bm{\theta}) )
\end{equation}
Here, $\bm{J}_{v}$ and $\bm{J}_{\omega}$ are Jacobian matrices of relative position and orientation of the end-effector with respect to the multirotor base. $\mu_v$ and $\mu_\omega$ are positive weights. A higher value of $\phi_m$ indicates greater manipulability, signifying that the robot arm configuration is positioned farther away from singularity \cite{planner_manipulability}.
Finally, denoting $\bm{x}_d := (\bm{p}_d, \bm{R}_d, \bm{\theta}_d) \in \mathbb{R}^3 \times \mathsf{SO}(3) \times \mathbb{R}^n$, ${}^E\bm{x}_d := ({}^E\bm{p}_d, {}^E\bm{R}_d) \in \mathbb{R}^3 \times \mathsf{SO}(3)$ and $\bm{u}_d := [\bm{v}_d;\bm{\omega}_d; \dot{\bm{\theta}}_d] \in \mathbb{R}^{6+n}$, we design the following state cost $\phi_{x}$ and input cost $\phi_{u}$
\begin{equation}  \label{planner: WBcost}
\begin{aligned}
    \phi_{x}(\bm{x}_d, {}^E\bm{x}_d) := &\lVert {}^E \bm{p}(\bm{x}_d) - {}^E \bm{p}_d\rVert_{\bm{Q}_p}^{2} - \phi_m(\bm{\theta}_d)\\
                                        & + \text{tr}({\bm{Q}_R}(\bm{I}_3- {}^E \bm{R}_d^\top {}^E\bm{R}(\bm{x}_d)))  \\
    \phi_{u}(\bm{u}_d) := &\lVert \bm{u}_d \rVert_{\bm{R}_u}^{2},
\end{aligned}
\end{equation}
where $\bm{Q}_p$, $\bm{Q}_R$ and $\bm{R}_u$ are all positive definite weight matrices. ${}^E \bm{p}(\bm{x}_d)$ and ${}^E\bm{R}(\bm{x}_d)$ are computed by forward kinematics using $\bm{x}_d$.

\subsubsection{Kinematics-level nonlinear model predictive control}
With the obstacle avoidance constraints (\ref{planner: WBCollisionAvoidance}) and the cost functions (\ref{planner: WBcost}), we formulate and solve the following OCP at each replanning interval, employing a nonlinear model predictive control (NMPC) approach:
\begin{equation} \label{planner: WBNMPCproblem}
\begin{aligned}
\min_{\bm{x}_d, \bm{u}_d} & \phi_{x}(\bm{x}_d(N_H), {}^E\bm{x}_d(N_H))\\
                          & + \sum_{k=0}^{N_H-1} { \phi_{x}(\bm{x}_d(k), {}^E\bm{x}_d(k)) + \phi_{u}(\bm{u}_d(k)) } \\
\textrm{s.t.} \quad  &\bm{x}_d(0) = (\bm{p},\bm{R},\bm{\theta}) \\
     \forall_k \quad &\bm{x}_d(k+1) = \bm{f}_{x}(\bm{x}_d(k), \bm{u}_d(k)) \\
     \forall_k \quad &\bm{u}_m \preceq \bm{u}_d(k) \preceq \bm{u}_M \\
     \forall_k \quad &\bm{A}_{\theta}\bm{\theta}_d(k) \preceq \bm{b}_{\theta} \\
     \forall_{i,j,k} \quad &\hat{h}(\mathcal{A}_i(k), \mathcal{O}_j) > 0  
\end{aligned}
\end{equation}
Here, $\bm{p}, \bm{R}$ and $\bm{\theta}$ are the measured values at every replanning time. 
$\bm{f}_{x}$ represents discretized kinematic relations for the position, rotation matrix, and joint angle. See Appendix \ref{appendix: def of online kinematics} for the detail.
The upper and lower bounds for input $\bm{u}_d$ are denoted as $\bm{u}_M$ and $\bm{u}_m$. The fourth inequality constraint, i.e. $\bm{A}_{\theta}\bm{\theta}_d(k) \preceq \bm{b}_{\theta}$, prevents self-collision between the manipulator and the multirotor base, where $\preceq$ denotes element-wise inequality. \blue{The horizon length $N_H$ is $T_H/\Delta t$.}
After solving the optimization problem (\ref{planner: WBNMPCproblem}) and refining the whole-body trajectory, we utilize $\bm{p}_d$, $\bm{v}_d$, $\bm{R}_d$, and $\bm{\omega}_d$  as desired values for the controller in (\ref{eq:ctrller - translation}) and (\ref{eq:ctrller - rotation}). Simultaneously, $\bm{\theta}_d$ and $\dot{\bm{\theta}}_d$ serve as desired values for the external robot arm controller.

    In scenarios where the manipulator is constrained to planar motion, the manipulability index $\phi_m$ in (\ref{planner: WBManipulabilityIndex}) becomes 0 as $\bm{J}_v$ and $\bm{J}_\omega$ are not full rank.
    In such instances, $\phi_m$ can be computed utilizing the Jacobian matrices projected onto the plane of motion.
    Moreover, if the manipulator comprises solely revolute joints rotating in the same direction, the term associated with $\bm{J}_{\omega}$ remains constant and can be excluded from the cost function.
    For the case when the goal end-effector orientation is deemed inconsequential and omitted to be solved in (\ref{planner: EEOCPRotation}),
    excluding the term $
    \text{tr}({\bm{Q}_R}(\bm{I}_3- {}^E \bm{R}_d^\top {}^E\bm{R}(\bm{x}_d)))$ in (\ref{planner: WBcost}) suffices.
    This is because the remaining state cost also contains all of the state variable $\bm{x}_d$.

\section{Experimental results}

\subsection{Setup}

\begin{table} 
\scriptsize
\centering
\caption{Main components of the system}
\label{tb:components}
\begin{tabular}{@{}lll@{}} 
\toprule
Component & Product name  & Quantity\\ 
\midrule
\begin{tabular}[c]{@{}l@{}} Onboard computer \\ Rotor \\ Propeller \\ Servo (base) \\ Servo (manipulator) \\ ESC \\ PWM generator \\ IMU sensor \\ Battery
\end{tabular} 
& 
\begin{tabular}[c]{@{}l@{}} Intel NUC i3 \\ Armattan Oomph TITAN 2306/2450KV \\ APC BD6x4.2E-3-B4 \\ Dynamixel XC330 series \\ Dynamixel XM430 \& XC330 series \\ Hobbywing XRotor Micro 40A 4in1 \\ Nucleo F446RE \\ Vectornav VN-100 \\ Turnigy 4200 mAh 4S LiPo \end{tabular}
& 
\begin{tabular}[c]{@{}c@{}} 1 \\ 6 \\ 6 \\ 6 \\ 4 \\ 2 \\ 1 \\ 1 \\ 1 \end{tabular} \\
\bottomrule
\end{tabular}\\
\end{table}


We construct a customized OAM whose CAD model can be found in Fig. \ref{fig:control_allocation}. Table \ref{tb:components} lists main components of the platform whose net weight is about $2.13$ \si{kg}. The manipulator is composed of $4$ revolute joints where the first three joints rotate in the same direction while the last joint is for a gripper. Algorithms including the proposed controller, planner, and state estimator run in an onboard computer where we use Robot Operating System (ROS) in Ubuntu 20.04. The current state of the robot is estimated by a state estimation algorithm \cite{state-estimator} which executes sensor fusion for measurements from Optitrack motion capture system and inertial measurement unit (IMU). 
The overall algorithm for hardware experiments is visualized in Fig. \ref{fig:ExperimentsFlowDiagram-all}.

Three experiments are conducted. In the first experiment, we show the performance of the proposed controller by comparing the result that is obtained with baseline controllers. The objective is to regulate the pose of the multirotor base in the presence of the robotic arm's motion, and better regulation performance can be observed with the proposed control law. The second and third experiments are to demonstrate the proposed framework in one precise manipulation task of grasping-and-pulling an object. We consider two different environment settings: grasping-and-pulling an object 1) on the ground and 2) on a table. Compared to a conventional aerial manipulator based on an underactuated multirotor base inhering a limited workspace, the OAM equipped with the proposed framework could accomplish the task in both environments by leveraging omnidirectionality and the extended workspace. 

\subsection{Implementation details}

\begin{table} 
\footnotesize
\centering
\caption{Controller and control allocation parameters \\ (diagonal elements for matrices)}
\label{tb:parameters/gains}
\begin{tabular}{@{}l l | l l@{}} 
\toprule
\multicolumn{4}{c}{Controller} \\
\midrule
Parameter & Value  & Parameter & Value\\ 
\midrule
    \begin{tabular}{@{}l@{}} $\bar{m}$ [$\si{kg}$] \\ $\bm{K}_{tp}$ \\ $\bm{K}_{td}$ \\ $\bm{K}_{ti}$ \\ $\bm{\Lambda}_t$ \\ $\bm{\Gamma}_t$ \\ $\bm{\Theta}_t$ \\ $\rho_t$ \end{tabular} 
&
    \begin{tabular}{@{}l@{}} $2.13$ \\ $(8,8,8)$ \\ $(5,5,5)$ \\ $(2,2,4)$ \\ $(3,2,2)$ \\ $(2,2,2)$ \\ $(3,3,3)$ \\ $1$ \end{tabular}
&
    \begin{tabular}{@{}l@{}} $\bar{\bm{J}}_b$ [$\si{kgm^2}$]\\ $\bm{K}_{rp}$ \\ $\bm{K}_{rd}$ \\ $\bm{K}_{ri}$ \\ $\bm{\Lambda}_r$ \\ $\bm{\Gamma}_r$ \\ $\bm{\Theta}_r$ \\ $\rho_r$ \end{tabular}
& 
    \begin{tabular}{@{}l@{}} $(0.02,0.025,0.035)$ \\ $(15,20,10)$ \\ $(10,9,5)$ \\ $(0.08,0.08,0.08)$ \\ $(8,8,8)$ \\ $(0.2,0.2,0.2)$ \\ $(10,10,10)$ \\$0.02$  \end{tabular} \\
\midrule
\multicolumn{4}{c}{Allocation} \\
\midrule
\multicolumn{2}{@{}l@{}}{Parameter} & \multicolumn{2}{@{}l@{}}{Value} \\
\midrule
\multicolumn{2}{@{}l@{}}{$\bm{W}$} & \multicolumn{2}{@{}l@{}}{$(1,1,0.6,0.6,1,1,1,1,0.6,0.6,1,1)$} \\
\bottomrule
\end{tabular}\\
\end{table}

\begin{table}
\footnotesize
\centering
\caption{Planner parameters and numerical integration method \\ (block diagonal matrices for $\bm{R}_u$)}
\label{table:planner_params}
\begin{tabular}{@{}l l | l l@{}} 
\toprule
\multicolumn{4}{c}{Offline} \\
\midrule
Parameter & Value  & Parameter & Value\\ 
\midrule
    \begin{tabular}{@{}l@{}} $\bm{R}_v, \bm{R}_\omega$ \\ $T_f$ [$\si{sec}$] \end{tabular} 
&
    \begin{tabular}{@{}l@{}} $\bm{I}_3$ \\ $15$ \end{tabular}
&
    \begin{tabular}{@{}l@{}} $\gamma_i$\\ $\Delta t$ [$\si{sec}$] \end{tabular}
& 
    \begin{tabular}{@{}l@{}} $3$ \\ $0.1$ \end{tabular} \\
\midrule
\multicolumn{4}{c}{Online} \\
\midrule
Parameter & Value  & Parameter & Value\\ 
\midrule
    \begin{tabular}{@{}l@{}} $\mu_v$ \\ $\bm{Q}_p$ \\ $\bm{Q}_R$ \\ \blue{$T_H$ [$\si{sec}$]} \end{tabular} 
&
    \begin{tabular}{@{}l@{}} $0.01$ \\ $5\bm{I}_3$ \\ $4\bm{I}_3$ \\ \blue{$1.5$} \end{tabular} 
&
    \begin{tabular}{@{}l@{}} $\mu_\omega$ \\ $\bm{R}_u$ \\ $\bm{u}_M, -\bm{u}_m$ \\ $\Delta t$ [$\si{sec}$] \end{tabular} 
& 
    \begin{tabular}{@{}l@{}} $-$ \\ $(0.01\bm{I}_3, 0.01\bm{I}_3, 0.1\bm{I}_3)$ \\ $[\bm{1}_3; 0.5\pi\bm{1}_3; 0.25\pi\bm{1}_3]$ \\ $0.1$ \end{tabular}  \\
\midrule
\multicolumn{4}{c}{Numerical integration} \\
\midrule
Parameter & Method  & Parameter & Method\\ 
\midrule
${}^{E}\bm{f}_p, {}^{E}\bm{f}_R$ & RK4 & $\bm{f}_x$ & Euler \\
\bottomrule
{\footnotesize $\bm{1}_3 := [1;1;1]$}\\
\end{tabular}\\
\end{table}

Control gains we used during the whole experiments are listed in Table \ref{tb:parameters/gains}. In the gain tuning process, we first tune PID gains $\bm{K}_{tp},\bm{K}_{td},\bm{K}_{ti},\bm{K}_{rp},\bm{K}_{rd},\bm{K}_{ri}$ while setting all the other gains to be zero. After finding proper PID gains, we freeze the PID gains and tune the other gains related to the integral of the tanh of the error, which are $\bm{\Lambda}_t, \bm{\Gamma}_t, \bm{\Theta}_t, \rho_t, \bm{\Lambda}_r, \bm{\Gamma}_r, \bm{\Theta}_r, \rho_r$. We first set $\rho_t,\bm{\Gamma}_t$ and $\rho_r,\bm{\Gamma}_r$ to be the same scale as $\bm{K}_{ti}$ and $\bm{K}_{ri}$, respectively, and adjust them afterwards. Similarly, we initially take $\bm{\Lambda}_t, \bm{\Lambda}_r$ to be slightly larger than $1$ as they indicate the decay rate and tune them afterwards. Lastly, we initially set $\bm{\Theta}_t,\bm{\Theta}_r$ to be $1$ and increase them until achieving sufficient performance in disturbance rejection since larger $\bm{\Theta}_t,\bm{\Theta}_r$ \blue{are shown to be} effective in reducing the error bound. 

We empirically find that the $2^{nd}$ and $5^{th}$ rotors depicted in Fig. \ref{fig:control_allocation} go near saturation when the robot hovers at $90^\circ$ pitch angle with the identity weight matrix $\bm{W}$ in control allocation. This is because when hovering at $90^\circ$ pitch angle, 1) the pseudo-inverse solution finds the minimum norm solution and 2) other rotors are less effective in compensating gravity as they are tilted with respect to the gravity direction. Considering the robotic arm configuration in Fig. \ref{fig:control_allocation}, manipulation while maintaining non-zero or even $90^{\circ}$ pitch angle can frequently occur during experiments. To relieve this issue, we allocate values smaller than 1 to the elements related to the $2^{nd}$ and $5^{th}$ rotors in the weight matrix $\bm{W}$ in (\ref{eq: control allocation - pseudo inverse}). The value we use during all experiments can be found in Table \ref{tb:parameters/gains}.

Our experiments involve grasping a target object. Thus, a certain criterion is needed to determine whether the object is firmly grasped or not. For the criterion, we choose the current of the gripper servomotor. Before the experiments, the threshold for successful grasping is determined by securing various test objects with the gripper and observing the required value for it.
In actual experiments, a low-pass filter is applied to the gripper servomotor's current.
If the filtered value surpasses the pre-determined threshold after 1.5 seconds, the grasping is considered successful, and a new trajectory is planned for the subsequent mission, which is pulling.
The command to grasp is given when the position error between the end-effector and the goal is less than $3$ \si{cm}.
If the criterion is not met after the command, the gripper is reopened. This process is repeated until successful grasping occurs.

The collision avoidance with the ground is efficiently addressed by representing both the base and individual links of the OAM as spheres.
This pragmatic simplification allows us to express the constraint in the form of $r_i \leq z_i(t)$ where $r_i$ is the radius and $z_i$ is the height of the $i^{th}$ component of the OAM.
For OCPs described in (\ref{planner: EEOCPTranslation}), (\ref{planner: EEOCPRotation}), and (\ref{planner: WBNMPCproblem}), we employed the open-source tool CasADi \cite{CasADi}.
The chosen nonlinear optimization solver was IPOPT with HSL libraries \cite{HSLlib} serving as a linear solver. 
With this setup, the entire offline OCPs in (\ref{planner: EEOCPTranslation}) and (\ref{planner: EEOCPRotation}) are solved within 3-5 seconds.
The parameters used for our whole-body motion planner algorithm are shown in Table \ref{table:planner_params}.
Our OAM's revolute joints related to the motion of the manipulator rotate along the same direction, meaning that $\bm{J}_\omega$ becomes constant.
Thus, in the table, the value of $\mu_\omega$ is not listed. 

\subsection{Experiment 1: controller comparison}
\begin{figure*}[t]
    \centering 
    \includegraphics[width=0.3\linewidth]{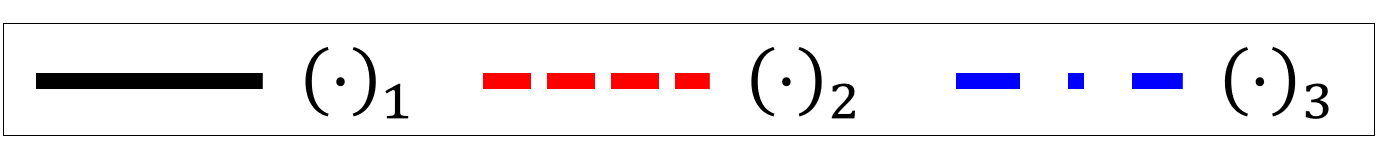}

    \begin{subcaptionbox}{Results of setting 1: pose regulation at $0^\circ$ pitch angle.\label{fig:ctrller-test-pitch0}}[0.49\linewidth]
        {\centering\includegraphics[width=\linewidth, trim={0cm 0cm 0cm 0cm}]{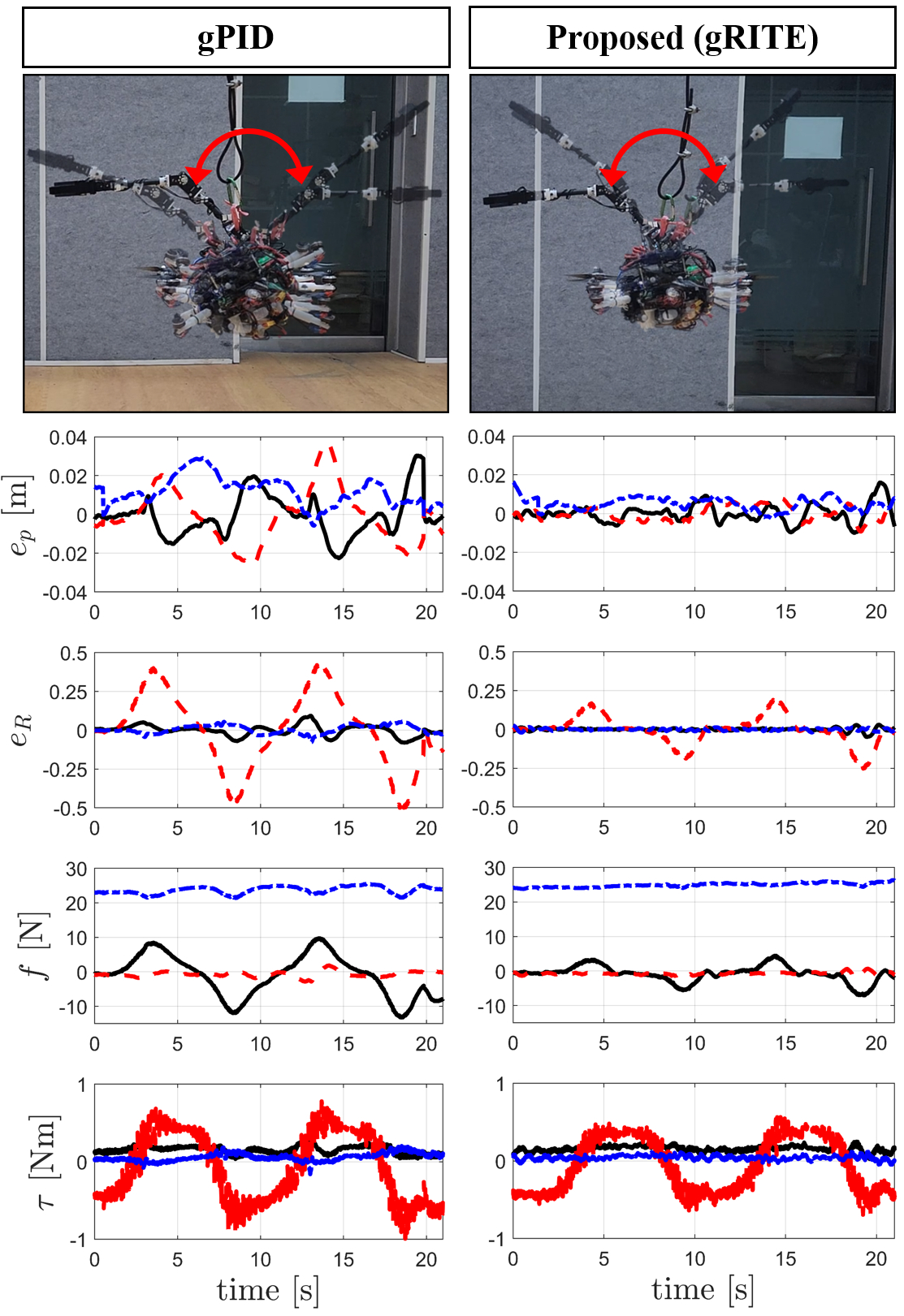}
        }
    \end{subcaptionbox}
    \hfill
    \begin{subcaptionbox}{Results of setting 2: pose regulation at $-30^\circ$ pitch angle.\label{fig:ctrller-test-pitch-30}}[0.49\linewidth]
        {\centering\includegraphics[width=\linewidth, trim={0cm 0cm 0cm 0cm}]{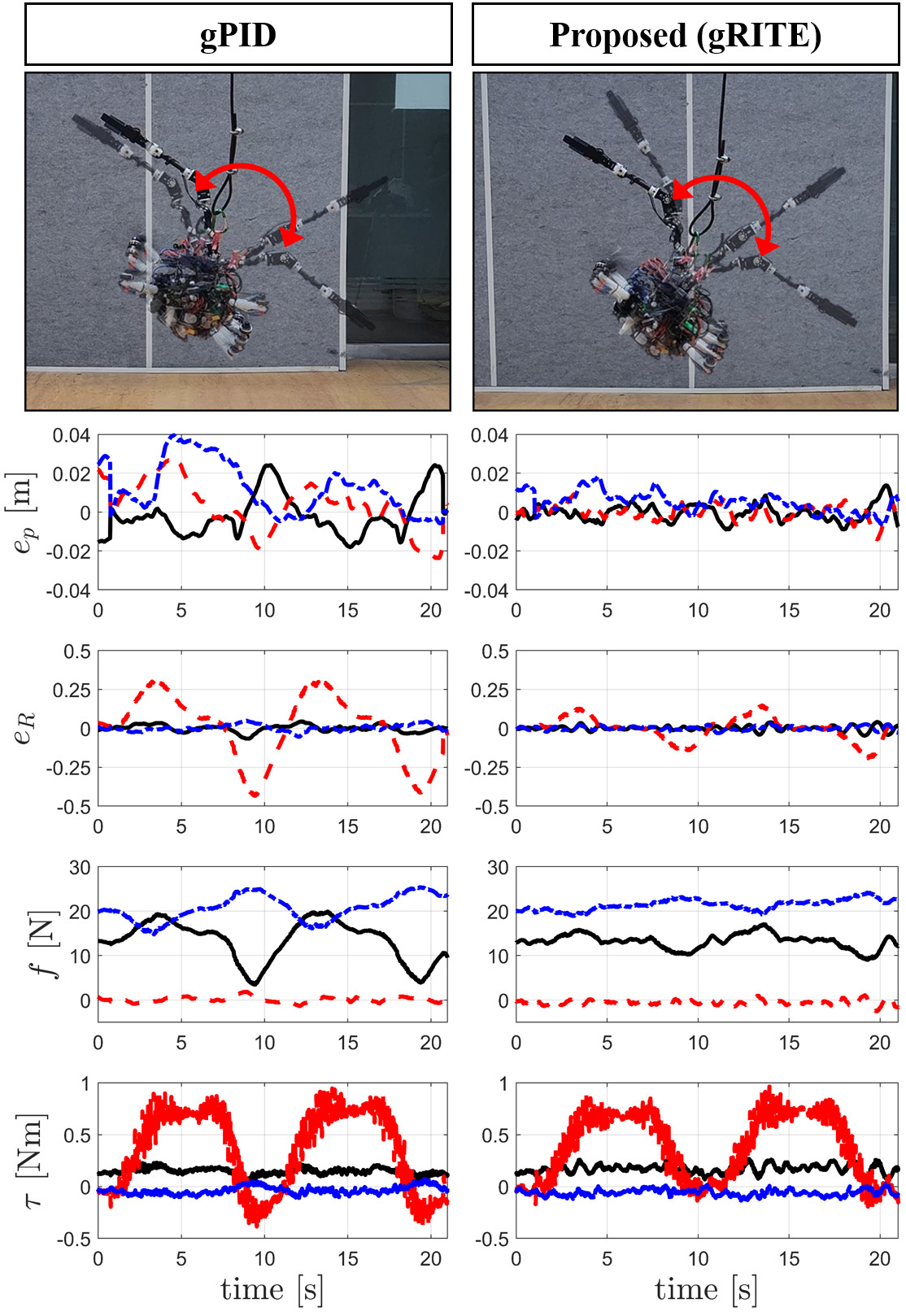}
        }
    \end{subcaptionbox}
    \caption{Results of Experiment 1: comparison between gPID and the proposed controller (gRITE).}
    \label{fig:ctrller-comp-gPID}
\end{figure*}

\begin{table*}[t]
\centering
\caption{\blue{Performance metrics of Experiment 1}}
\label{tb:exp1_metrics}
\resizebox{\textwidth}{!}{
\begin{tabular}{>{\centering\arraybackslash}p{0.09\linewidth}|
                >{\centering\arraybackslash}p{0.10\linewidth}| 
                >{\centering\arraybackslash}p{0.07\linewidth}
                >{\centering\arraybackslash}p{0.07\linewidth}
                >{\centering\arraybackslash}p{0.07\linewidth}
                >{\centering\arraybackslash}p{0.07\linewidth}
                >{\centering\arraybackslash}p{0.07\linewidth}|
                >{\centering\arraybackslash}p{0.07\linewidth}
                >{\centering\arraybackslash}p{0.07\linewidth}
                >{\centering\arraybackslash}p{0.07\linewidth}
                >{\centering\arraybackslash}p{0.07\linewidth}
                >{\centering\arraybackslash}p{0.07\linewidth}}

\toprule
 &        & \multicolumn{5}{c|}{setting 1 ($0^\circ$ pitch angle)} & \multicolumn{5}{c}{setting 2 ($-30^\circ$ pitch angle)} \\
 &        & gPID & gRISE & DOB & g$\mathcal{L}_1$ & proposed & gPID & gRISE & DOB & g$\mathcal{L}_1$ & proposed \\ 
\midrule
\multirow{3}{*}{\begin{tabular}[c]{@{}c@{}}Position \\ tracking\\ error${}^*$ \end{tabular}} 
 & RMS [cm]    & 1.31 & 1.04 & 1.59 & \textbf{0.451} & 0.488 & 1.44 & 0.786 & 1.76 & 0.723 & \textbf{0.518} \\
 & Mean [cm]   & 1.19 & 0.934 & 1.43 & \textbf{0.418} & 0.462 & 1.29 & 0.731 & 1.61 & 0.657 & \textbf{0.471} \\
 & Std [cm]    & 0.537 & 0.442 & 0.702 & 0.171 & \textbf{0.156} & 0.623 & 0.289 & 0.712 & 0.303 & \textbf{0.216} \\
\midrule
\multirow{3}{*}{\begin{tabular}[c]{@{}c@{}}Orientation \\ tracking\\ error${}^\dagger$ \end{tabular}} 
 & RMS [deg]    & 14.5 & 6.27 & 5.44 & 5.69 & \textbf{5.49} & 12.5 & 6.23 & 6.54 & 5.79 & \textbf{4.53} \\
 & Mean [deg]   & 11.6 & 5.29 & 4.50 & 4.78 & \textbf{3.89} & 10.2 & 5.11 & 5.82 & 4.59 & \textbf{3.48} \\
 & Std [deg]    & 8.64 & 3.36 & \textbf{3.06} & 3.08 & 3.87 & 7.13 & 3.56 & 2.99 & 3.54 & \textbf{2.91} \\
\bottomrule
\multicolumn{12}{l}{\footnotesize \textbf{Bold} numbers indicate the best performance (i.e., lowest error) among all controllers for each metric.} \\
\multicolumn{12}{l}{\footnotesize ${}^*$: Euclidean norm of $e_p$} \\
\multicolumn{12}{l}{\footnotesize ${}^\dagger$: geodesic distance $d_g$ between the two rotation matrices $R$ and $R_d$, i.e., $d_g = \cos^{-1}\left(\frac{\text{trace}(R^\top R_d) - 1}{2}\right)$}

\end{tabular}
}
\end{table*}

\begin{figure*}[t]
    \centering 
    \begin{subcaptionbox}{Results of setting 1: pose regulation at $0^\circ$ pitch angle.\label{fig:boxplot_exp1-setting1}}[0.49\linewidth]
        {\centering\includegraphics[width=\linewidth, trim={1cm 1.5cm 1.5cm 1cm}]{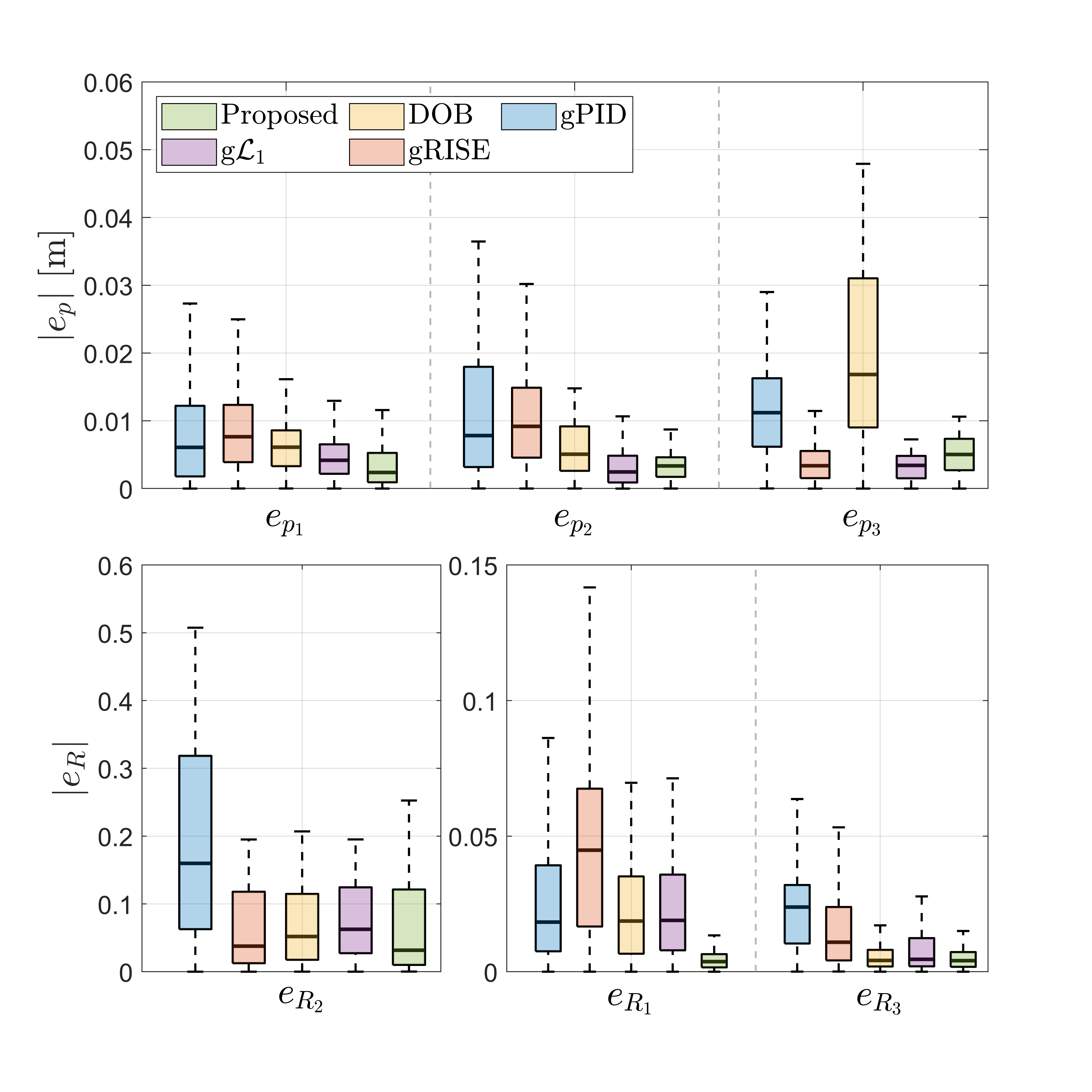}}
    \end{subcaptionbox}
    \hfill
    \begin{subcaptionbox}{Results of setting 2: pose regulation at $-30^\circ$ pitch angle.\label{fig:boxplot_exp1-setting2}}[0.49\linewidth]
        {\centering\includegraphics[width=\linewidth, trim={1cm 1.5cm 1.5cm 1cm}]{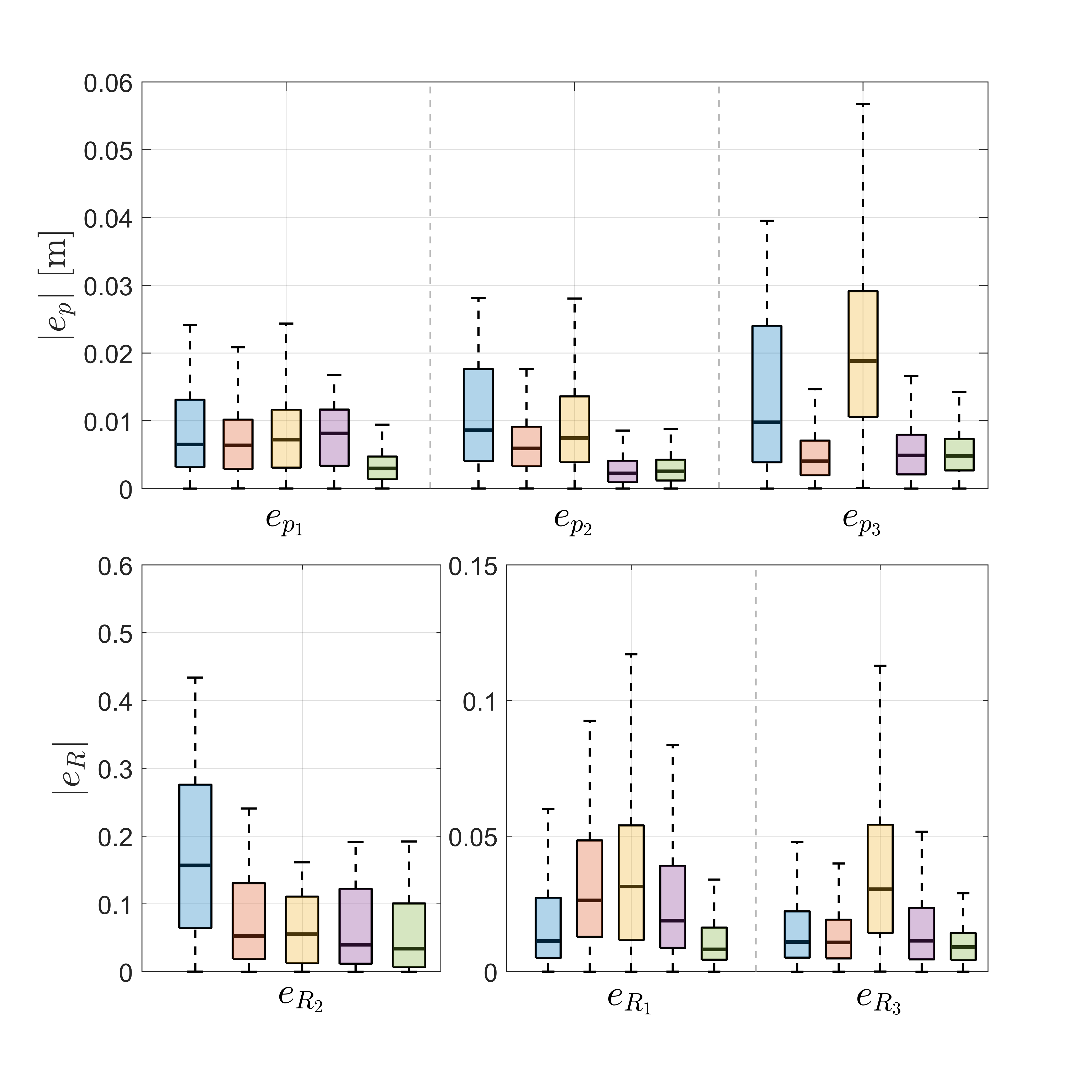}}
    \end{subcaptionbox}
    \caption{\blue{Box plot of Experiment 1 for comparison with baseline controllers.}}
    \label{fig:boxplot_exp1}
\end{figure*}

The first experiment is to show effectiveness of the proposed controller under disturbance, \blue{in comparison with the baseline controllers: 1) Euler-angle-based disturbance-observer (DOB) \cite{lee2021aerial} which is a nonlinear robust controller, 2) geometric PID controller (gPID) \cite{goodarzi2013geometric, planner_Omni_OfflineOnly}, 3) geometric $\mathcal{L}_1$ adaptive controller (g$\mathcal{L}_1$) \cite{wu2025L1quad}, and 4) geometric RISE controller (gRISE) \cite{gu2022agile}}. 
\blue{Unlike prior works on OAM that typically employ only gPID, our evaluation includes a broader range of robust and adaptive controllers, such as those tailored for multirotor platforms.} 

\blue{The baseline controllers share the same PD control structure, with an additional control term designed to reject disturbance. For fair comparison, we set the corresponding PD gains of the proposed controller to be identical to those used in the baseline controllers. 
Only the gains associated with the additional control terms were tuned.
Although g$\mathcal{L}_1$, DOB, and gRISE theoretically achieve better performance with higher gains, in real-world applications, measurement noise and physical limitations of actuators prevented arbitrarily increasing the gains. Accordingly, for each controller, the gains were tuned until a sufficiently low tracking error was achieved.}

The experiment is carried out in two different settings: pose regulation at $0^\circ$ pitch angle and $-30^\circ$ pitch angle. We intentionally oscillate the manipulator to provide external disturbance to the multirotor base. In all settings, only the first and second joints of the manipulator are commanded to move from $-45^\circ$ to $45^\circ$ with a period of $10$ \si{s}. 
\blue{As a representative example that most clearly shows the difference between controllers in the two settings, we first present comparison results between the gPID and the proposed controller in Figs. \ref{fig:ctrller-test-pitch0} and \ref{fig:ctrller-test-pitch-30}.} In both figures, the above composite images show the motions of the manipulator and the multirotor base, exhibiting the superior performance of the proposed controller. The below graphs in both figures display pose error $\bm{e}_p, \bm{e}_R$ and body force and torque $\bm{f},\bm{\tau}$ obtained during the experiments. 

In both experiments, while the maximum error in the translational motion does not exceed $0.02$ \si{m} in the results obtained with the proposed controller, that in the results obtained without the proposed controller nearly reaches $0.04$ \si{m}. The performance gap is more significant in the rotation direction which can be found in the second row of the graphs in Figs. \ref{fig:ctrller-test-pitch0} and \ref{fig:ctrller-test-pitch-30}. We presume that not the control law itself but the mechanical vibration of the manipulator and the resultant measurement noise in angular velocity are responsible for the jittering in $\tau_2$ during all experiments with or without the proposed control law. This is because such jittering does not occur in any other control inputs, and we have experienced mechanical vibration due to backlash and clearance of servomotors which may be improved by fabricating the OAM using high-end servomotors.

\blue{ 
The comparison results, including all baseline controllers, are summarized in Fig. \ref{fig:boxplot_exp1} and Table \ref{tb:exp1_metrics}.
Due to the motion of the robotic arm, the external disturbance is most significant along the body $y-\text{axis}$. Accordingly, the disturbance attenuation capability of each controller is most clearly reflected in the bottom-left plots of Figs. \ref{fig:boxplot_exp1-setting1} and \ref{fig:boxplot_exp1-setting2}.
The first, second, and third quartiles ($Q_1, Q_2, Q_3$) of the gPID controller are more than twice as large as those of the other controllers.
Thus, we can confirm that all other controllers are substantially robust to disturbance.
For DOB, which is designed based on Euler angles, both position and orientation errors noticeably increase when the pitch angle is $-30^\circ$ compared to the $0^\circ$ case.
This implies the importance of fully accounting for the nonlinear structure of $\mathsf{SO}(3)$ for accurate attitude control.
In the case of gRISE, the presence of the sign function leads to discontinuities in the first derivative of the control input.
As a result, its performance in directions that are relatively less influenced by external disturbances, namely position and the body $x- \text{and } z-\text{axes}$, is inferior to that of the other controllers, including gPID.
The $g\mathcal{L}_1$ controller employs a low-pass filter (LPF) to smooth out the rapidly varying control signal generated by its adaptation law.
This suppresses oscillatory inputs and improves tracking accuracy compared to other baseline controllers in practice.
As seen in the plots, it maintains relatively low position and orientation errors across most directions. This is best illustrated when the pitch angle is $0^\circ$, where it achieves the lowest RMS and mean position tracking errors among all controllers, as summarized in Table \ref{tb:exp1_metrics}.
While the LPF improves performance in the $0^\circ$ pitch case, its trade-off with performance leads to reduced accuracy at $-30^\circ$, where the gRITE controller outperforms it.
The proposed gRITE controller not only produces smooth control inputs via the tanh function, but also exhibits strong robustness against disturbances. The proposed controller consistently outperforms the baselines in almost all real-world experiments.
}

\subsection{Experiment 2: grasping-and-pulling an object on the ground}

\begin{figure*}[t!]
    \centering
    \includegraphics[width=0.5\linewidth]{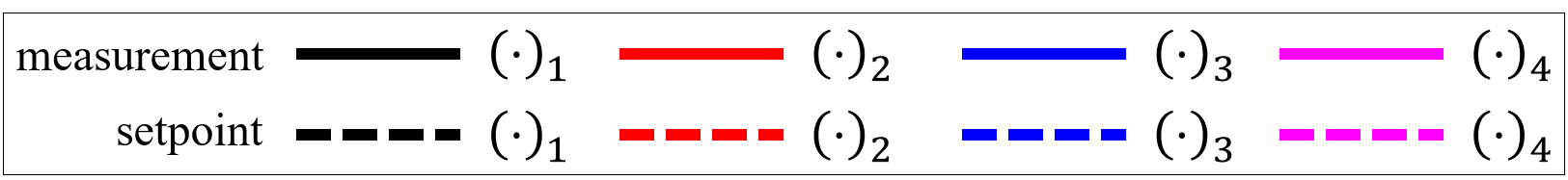}
    \includegraphics[width=1.0\linewidth]{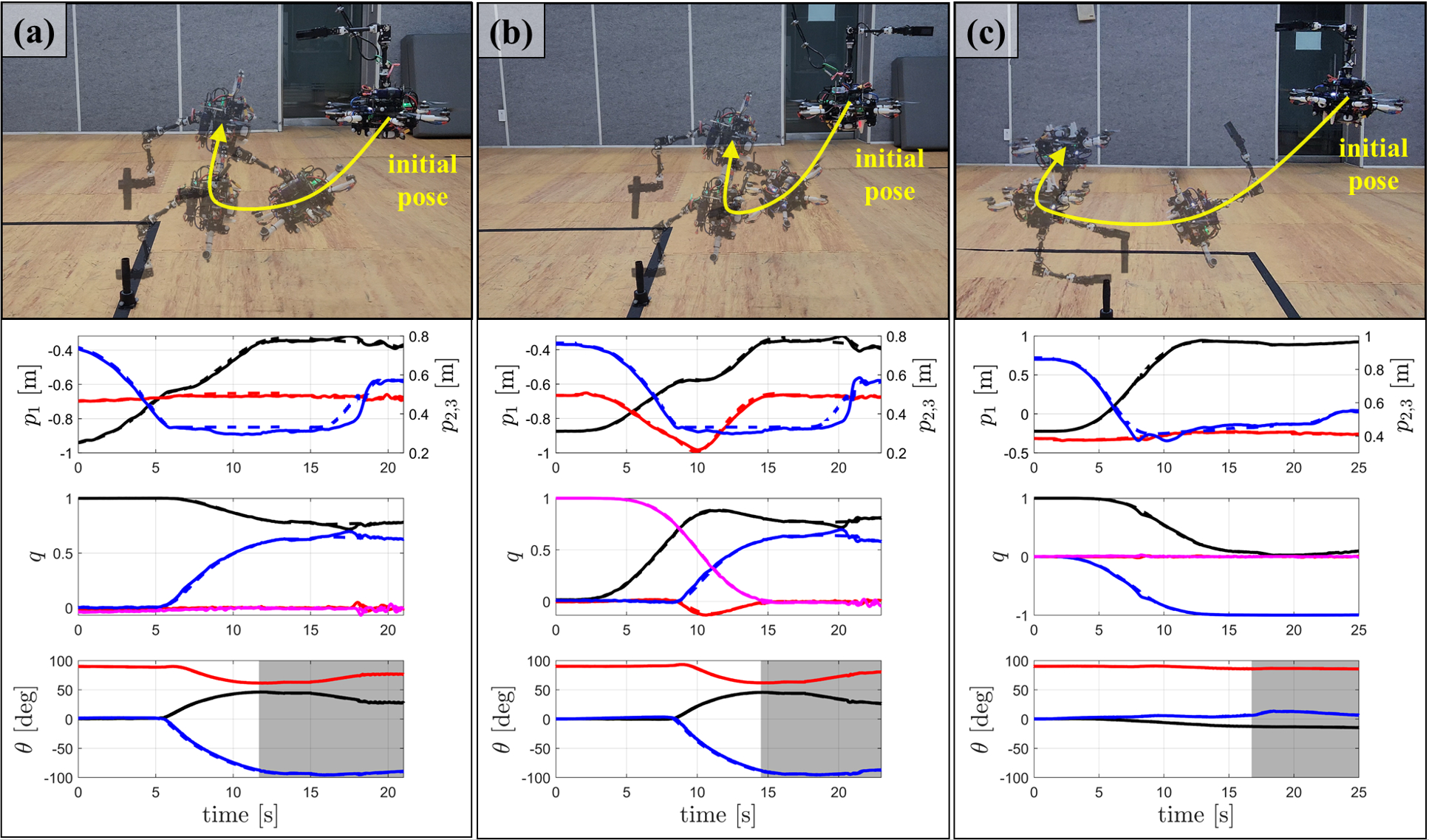}
    \caption{Results of Experiment 2: grasping-and-pulling an object on the ground. The \blue{notations} (a), (b) and (c) indicate to which scenario the figure corresponds to: (a) ground-basic, (b) ground-yaw, and (c) ground-pitch. We use quaternion $q = [q_w,q_x,q_y,q_z]$ to represent the orientation of the multirotor base.}
    \label{fig:ground-all}
\end{figure*}

\begin{figure}[h!]
    \centering
    \includegraphics[width=1.0\linewidth]{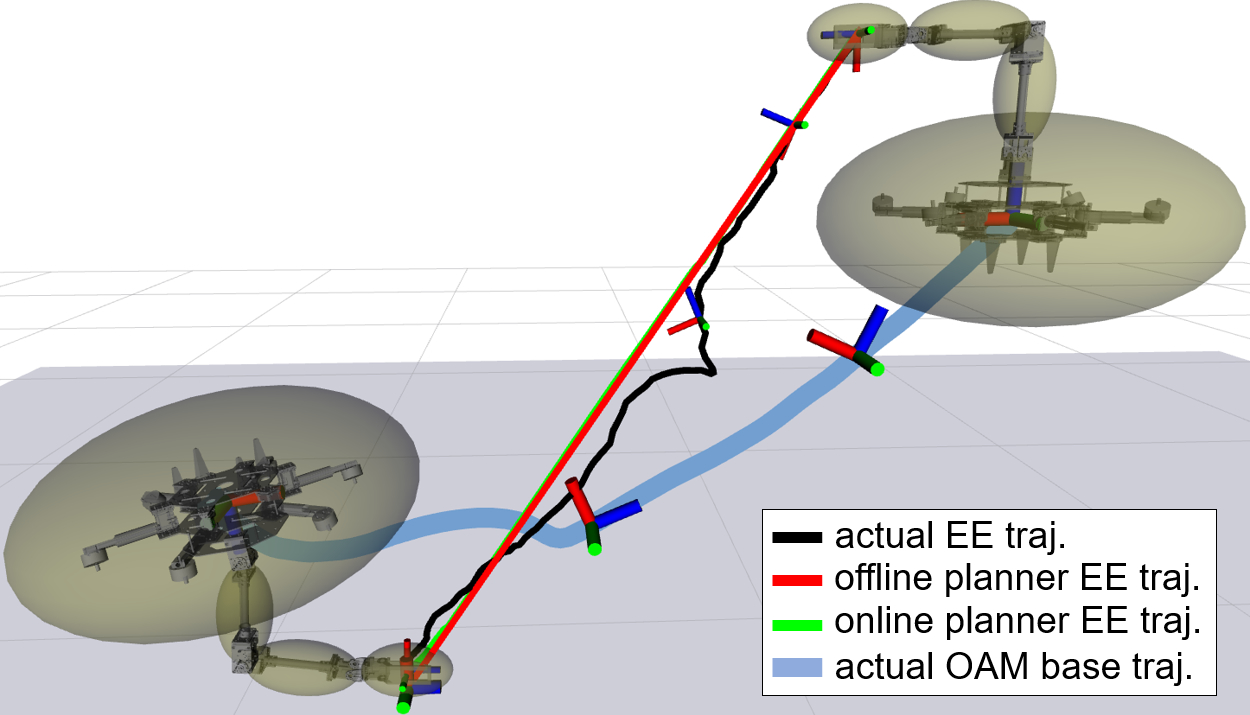}
    \caption{Visualization of trajectories computed by the whole-body motion planner in Experiment 2 setting (c) ground-pitch. The blue line indicates the motion of the multirotor base, and the red and green lines are end-effector trajectories computed by the offline and online planners. The black line is the actual trajectory traversed by the end-effector.}
    \label{fig:RVIZ_ground_pitch}
\end{figure}

\begin{figure}[h!]
    \centering
    \includegraphics[width=1.0\linewidth]{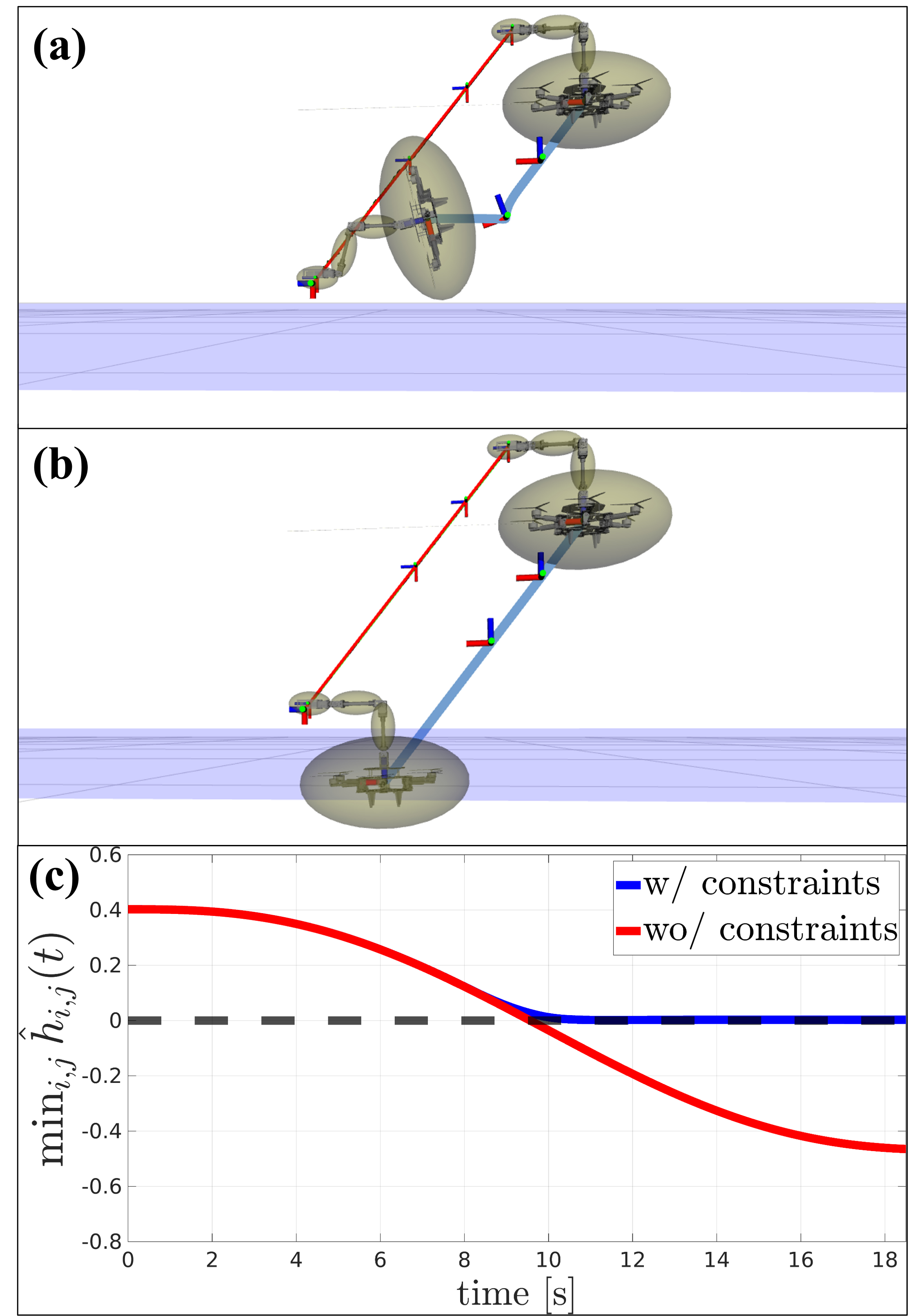}
    \caption{\blue{Simulated behaviors in the ground-basic scenario under the same initial hovering condition, with (a) and without (b) the collision avoidance constraint in (\ref{planner: WBCollisionAvoidance}). The minimum constraint at time $t$ $\hat{h}_{i,j}(t) = \hat{h}(\mathcal{A}_i(t), \mathcal{O}_j)$ is plotted in (c), while the translucent blue surface in (a) and (b) represents the ground.}}
    \label{fig:Comparison_onground_collision}
\end{figure}

The second experiment is grasping-and-pulling an object on the ground which can hardly be conducted with a conventional aerial manipulator whose manipulator is attached on the top as the OAM in Fig. \ref{fig:thumbnail}. We carry out three different scenarios which we call (a) ground-basic, (b) ground-yaw, and (c) ground-pitch. All scenarios and the data collected during the experiments are visualized in Fig. \ref{fig:ground-all}. Compared to (a) ground-basic, the initial orientation of the OAM in (b) ground-yaw is $180^\circ$ rotated in the yaw direction. (c) ground-pitch starts with the same orientation as (a) ground-basic, but the target end-effector orientation is $180^\circ$ rotated in the pitch direction. 

We could confirm that the proposed whole-body motion planning algorithm can compute a collision-free and goal-reaching trajectory in all three scenarios having either different initial conditions or different target poses. For the ground-pitch scenario, we visualize the end-effector trajectory computed by the first-step offline planner (red), the whole-body trajectory generated by the second-step online planner (green) and the actual traversed end-effector trajectory (black) in Fig. \ref{fig:RVIZ_ground_pitch}. Each ellipsoid represents each single rigid body comprising the OAM, used in deriving the collision avoidance constraints. We could also validate online replannability of the second-step NMPC-based planning algorithm in Table \ref{tb:exp23_metrics}. The maximum computation time is less than $20$ \si{ms} in all scenarios, indicating online replannability faster than $50$ \si{Hz}. 

The composite images at the top of Fig. \ref{fig:ground-all} show the motion of the OAM during experiments. Position $\bm{p}$ and orientation $\bm{q}$ of the multirotor base and joint angles of the manipulator $\bm{\theta}$ are visualized in the below graphs. As the multirotor rotates more than $90^\circ$ pitch angle, we use quaternion $q=[q_w,q_x,q_y,q_z]$ in representing the orientation. Thanks to the omnidirectionality of the OAM platform and the capability of the proposed framework to address such merit in both planning and control, the mobile manipulator successfully performs precise manipulation while enjoying the extended workspace. As described in the composite images of Fig. \ref{fig:ground-all}, the three experiments demonstrate precise manipulation of grasping-and-pulling while maintaining the pitch angle over $90^\circ$ and even near $180^\circ$.

{Quantitative results are summarized in Table \ref{tb:exp23_metrics}, and a detailed analysis of state errors and computation time is illustrated in the box plots in Fig. \ref{fig:boxplot_exp23}. All scenarios in experiment 2 exhibit position tracking errors in terms of both RMS and mean below $2$ \si{cm}, and orientation tracking errors remain within $3^\circ$. These errors are computed using the Euclidean norm of $e_p$ and the geodesic distance between the desired and actual rotation matrices. Fig. \ref{fig:boxplot_exp23} further provides component-wise distributions of the position and orientation errors. For each element of $e_p$ and $e_R$, the box plots capture the median and quartiles over time. This offers insight into axis-wise behavior of the tracking performance. For instance, the position error in $z$-direction $e_{p_3}$ tends to show slightly larger deviation across different scenarios, which can be attributed to the object not being instantly detached during the initial pulling motion.}

\blue{
To compare the overall behaviors with and without the collision avoidance constraint, simulation results for the ground-basic scenario are shown in Fig. \ref{fig:Comparison_onground_collision}. When the collision constraints (\ref{planner: WBCollisionAvoidance}) are not considered, the planned trajectory penetrates the ground, as shown in (b), and the constraint value drops significantly below zero, as seen in (c).
In contrast, as illustrated in (a), when the constraints are imposed, the planner avoids collision by tilting the OAM’s base toward the target, effectively utilizing its omnidirectional flight capability. In this case, we can guarantee collision avoidance, as the minimum value among the collision avoidance constraints remains above zero for all time, as shown in (c).
These results suggest that under the collision avoidance constraint, the planner can utilize the OAM’s omnidirectional flight capability to avoid collision and promote effective use of the extended workspace when reaching the target.}


\subsection{Experiment 3: grasping-and-pulling an object on a table}

\begin{figure*}[t!]
    \centering
    \includegraphics[width=0.5\linewidth]{jpg_figures/figure_6,9-legend.jpg}
    \includegraphics[width=0.75\linewidth]{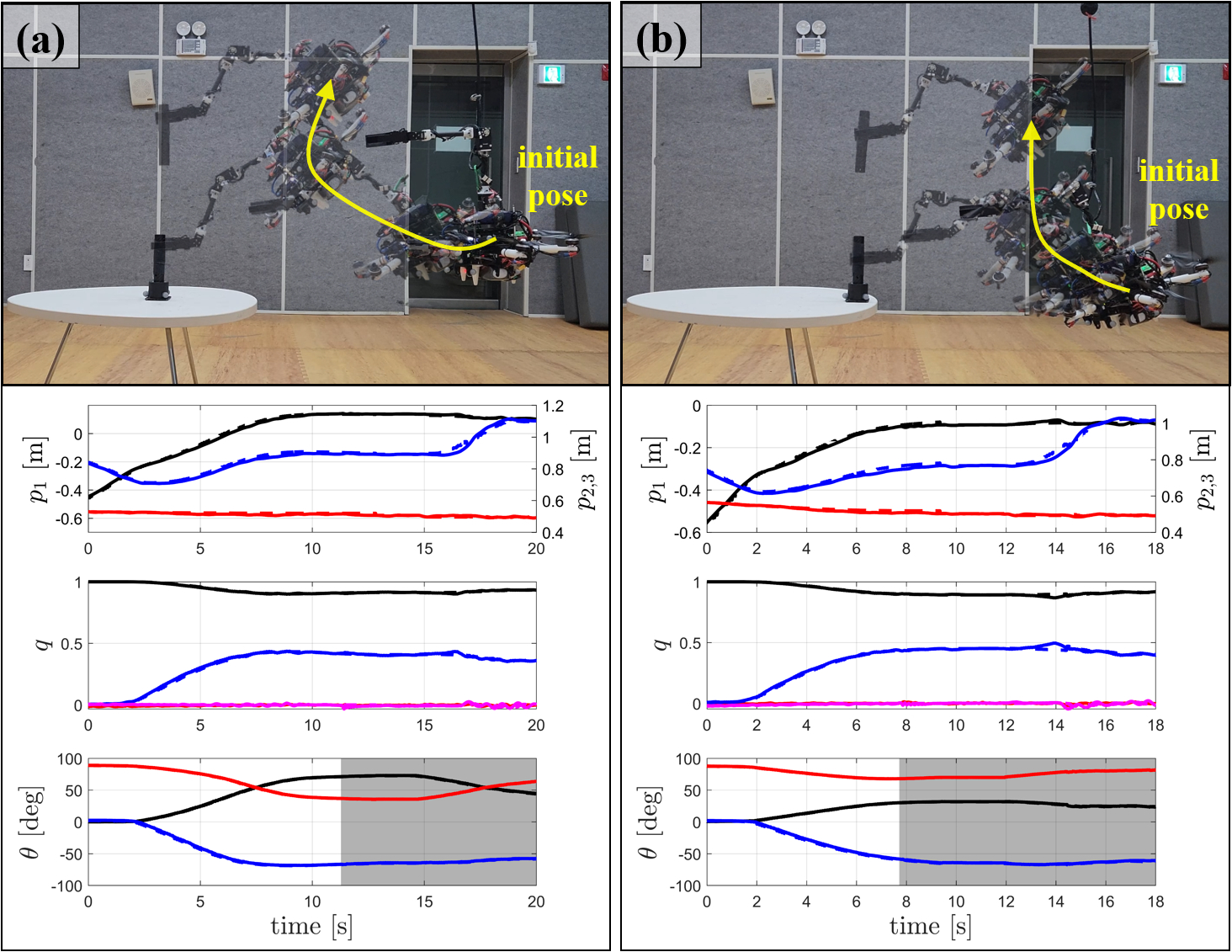}
    \caption{Results of Experiment 3: grasping-and-pulling an object on a table. The \blue{notations} (a) and (b) indicate to which scenario the figure corresponds to: (a) table-far, (b) table-close. We use quaternion $q = [q_w,q_x,q_y,q_z]$ to represent the orientation of the multirotor base.}
    \label{fig:table-all}
\end{figure*}

\begin{figure}[h!]
    \centering
    \includegraphics[width=1.0\linewidth]{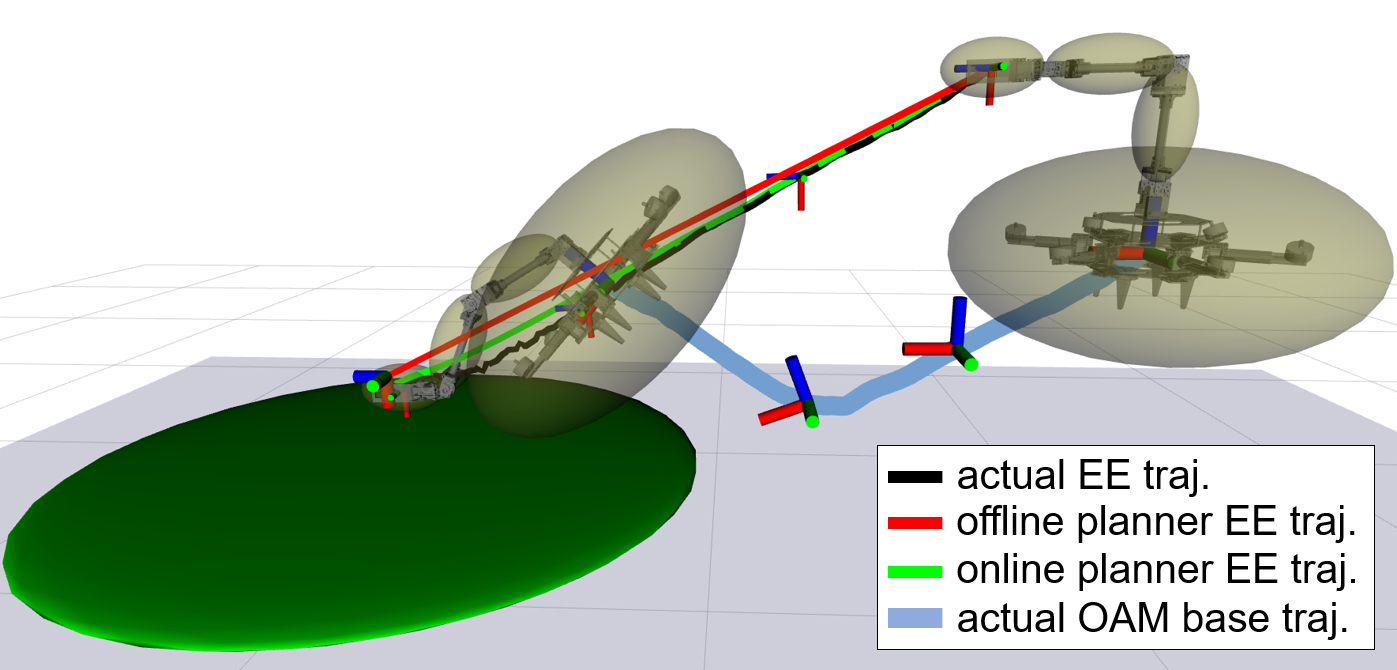}
    \caption{Visualization of trajectories computed by the whole-body motion planner in Experiment 3 setting (a) table-far. The blue line indicates the motion of the multirotor base, and the red and green lines are end-effector trajectories computed by the offline and online planner. The black line is the actual trajectory traversed by the end-effector.}
    \label{fig:RVIZ_Table_middle}
\end{figure}

\begin{figure}[h!]
    \centering
    \includegraphics[width=1.0\linewidth]{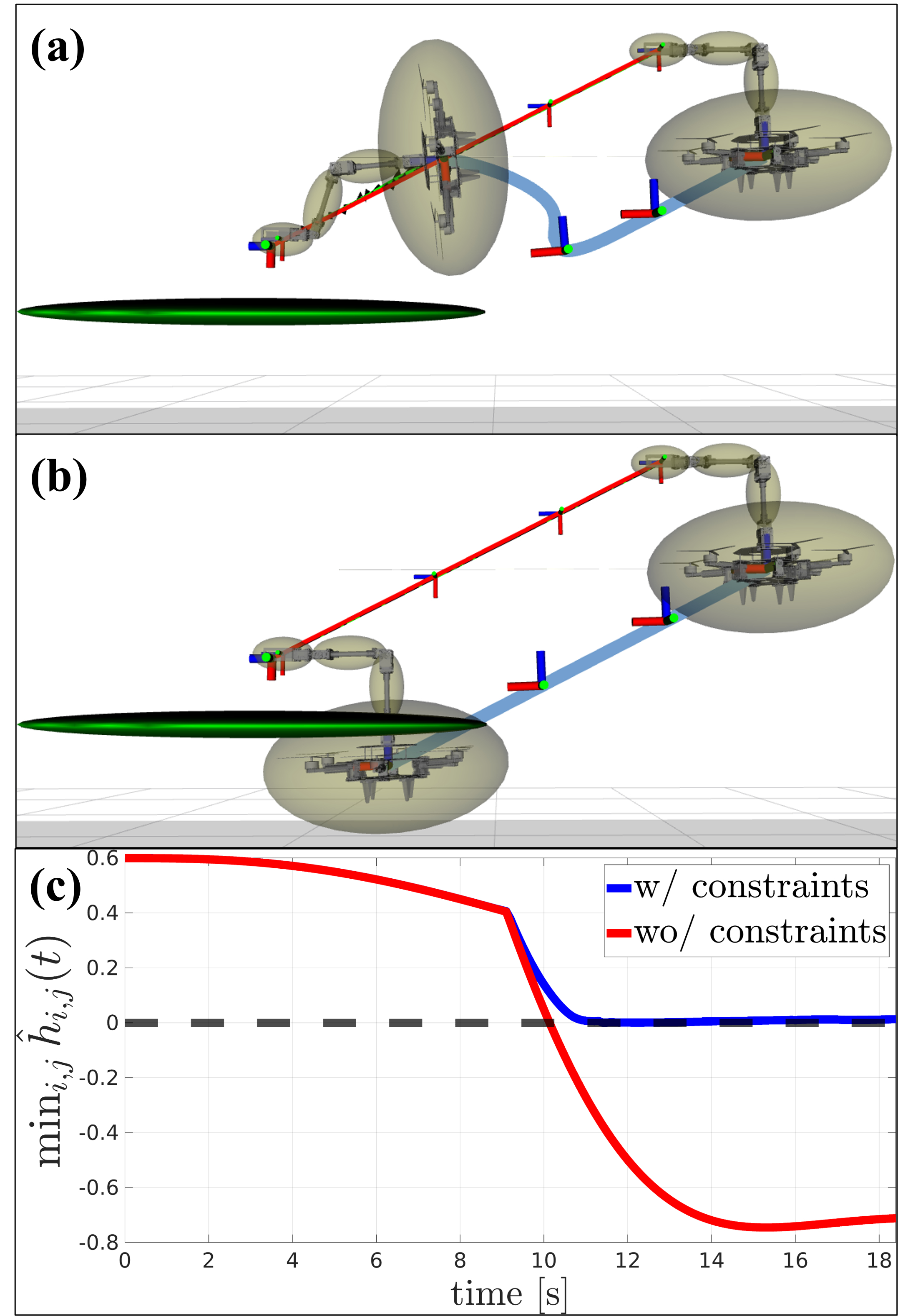}
    \caption{\blue{Simulated behaviors in the table-far scenario under the same initial hovering condition, with (a) and without (b) the collision avoidance constraint in (\ref{planner: WBCollisionAvoidance}). The minimum constraint at time $t$ $\hat{h}_{i,j}(t) = \hat{h}(\mathcal{A}_i(t), \mathcal{O}_j)$ is plotted in (c), while the green ellipsoid and the gray surface in (a) and (b) represent the table and the ground, respectively.}}
    \label{fig:Comparison_ontable_collision}
\end{figure}

The last experiment is to validate applicability of the proposed framework in a different setting of grasping-and-pulling an object on top of a table. The OAM should now additionally avoid collision with the table while accomplishing the task. Two different scenarios are considered: (a) table-far and (b) table-close. The results are summarized in Fig. \ref{fig:table-all}. The proposed whole-body motion planner computes a trajectory so that the OAM simultaneously tilts the pitch angle and stretches the manipulator to satisfy both the collision avoidance constraint and the goal-reaching objective. It is noticeable that when the object on the table is far from the robot (i.e. scenario (a)), to ensure safety, the proposed whole-body motion planner computes a trajectory not only to tilt the pitch angle and stretch the manipulator but also to stay above the table. 
Although a non-negligible ground effect from the table exists while the OAM being above the table in scenario (a), sufficient tracking performance to accomplish the precise manipulation task can be achieved with the proposed controller.

The computation time taken for solving the second step NMPC is listed in Table \ref{tb:exp23_metrics}. As an additional object to avoid exists compared to the Experiment 2, a little longer computation time is required, but still about $30$ \si{Hz} on average is obtained, which is sufficient for online replanning. Trajectories computed from the offline and online planning algorithms and the actual traversed trajectory are visualized in Fig. \ref{fig:RVIZ_Table_middle} for scenario (a).

\begin{figure*}[h]
    \centering
    \includegraphics[width=1.0\linewidth, trim={1.5cm 0 3cm 0}]{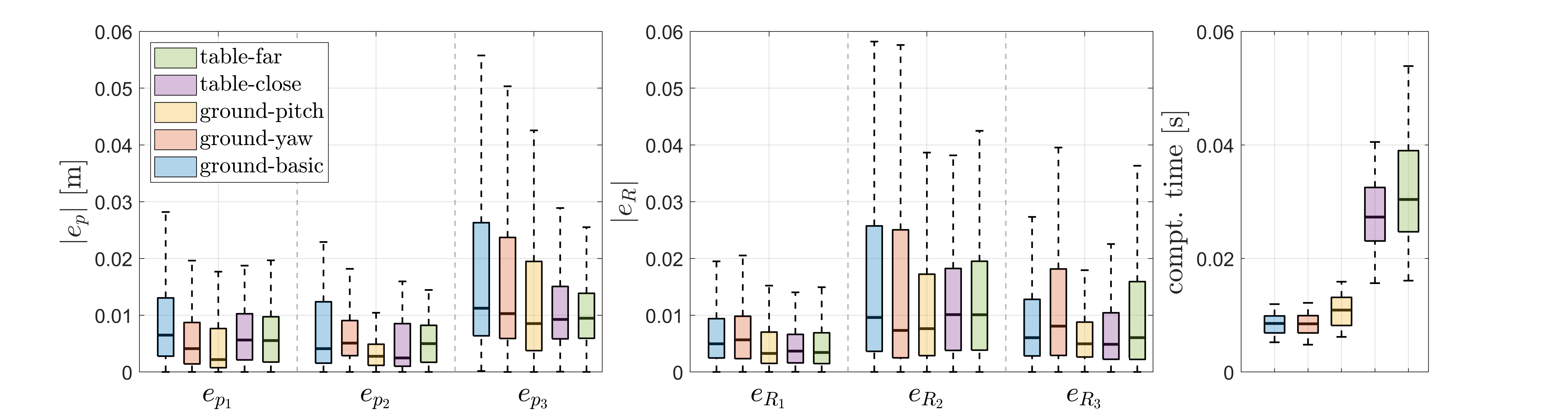}
    \caption{\blue{Box plot for state errors ($e_p$ and $e_R$) and NMPC computation time of Experiments 2 and 3.}}
    \label{fig:boxplot_exp23}
\end{figure*}

{Quantitative results are summarized in Table \ref{tb:exp23_metrics}, and a detailed analysis of state errors and computation time is illustrated in the box plots in Fig. \ref{fig:boxplot_exp23}. All scenarios in experiment 3 exhibit position tracking errors (in terms of both RMS and mean) below $1$ \si{cm}, and orientation tracking errors remain within $2^\circ$.} 
\blue{
Fig. \ref{fig:Comparison_ontable_collision} shows the effect of the collision avoidance constraint (\ref{planner: WBCollisionAvoidance}) in the table-far scenario. Similar to the ground-basic case, the OAM avoids collision with the table by ascending above it and simultaneously tilts its base to effectively reach the target located at the middle of the table.
}

\begin{table*}[t!]
\centering
\caption{\blue{Performance metrics of Experiments 2 and 3}} 
\small
\label{tb:exp23_metrics}
\begin{tabular}{c | c | c c c | c c}
\toprule
 &        & \multicolumn{3}{c|}{Experiment 2} & \multicolumn{2}{c}{Experiment 3} \\
 &        & ground-basic & ground-yaw & ground-pitch & table-far & table-close \\ 
\midrule
\multirow{3}{*}{\begin{tabular}[c]{@{}c@{}}Position \\ tracking\\ error${}^*$ \end{tabular}} 
 & RMS [cm]    & 1.84              & 1.69            & 1.29  & 0.955            & 0.960              \\
 & Mean [cm]   & 1.34   & 1.23  & 0.960  & 0.814   & 0.833               \\
 & Std [cm]    & 1.26   & 1.17  & 0.858  & 0.500   & 0.477               \\
\midrule
\multirow{3}{*}{\begin{tabular}[c]{@{}c@{}}Orientation \\ tracking\\ error${}^\dagger$ \end{tabular}} 
 & RMS [deg]    & 2.64              & 2.74      & 1.94  & 1.87        & 1.46             \\
 & Mean [deg]   & 1.82              & 1.92      & 1.18  & 1.31        & 1.23             \\
 & Std [deg]    & 1.91              & 1.96      & 1.54  & 1.34        & 0.788            \\
\midrule
\multirow{3}{*}{\begin{tabular}[c]{@{}c@{}}NMPC\\ computation \\ time \end{tabular}} 
 & Min [ms]    & 5.23             & 4.80           & 6.18    & 14.2 & 13.8         \\
 & Max [ms]    & 12.0             & 12.2           & 15.9    & 69.4 & 50.5            \\
 & Mean [ms]   & 8.50             & 8.42           & 10.7    & 27.8 & 29.2             \\
\bottomrule
\multicolumn{7}{l}{\footnotesize ${}^*$: Euclidean norm of $e_p$} \\
\multicolumn{7}{l}{\footnotesize ${}^\dagger$: geodesic distance $d_g$ between the two rotation matrices $R$ and $R_d$, i.e., $d_g = \cos^{-1}\left(\frac{\text{trace}(R^\top R_d) - 1}{2}\right)$}
\end{tabular}
\end{table*}

\section{Conclusion}
In this work, we presented a control and planning framework to enable mobile manipulation with arbitrary base position and orientation. To achieve this objective, we first constructed an omnidirectional aerial manipulator composed of an omnidirectional multirotor and a multi-DoF robotic arm. Then, a geometric robust controller is proposed for the multirotor base which we call a geometric robust integral of the tanh of the error (gRITE) controller. The controller is designed to ensure sufficient performance in the presence of external disturbance including aerodynamic effect, interaction wrench, and the uncertain motion of the robotic arm. The stability showed that the proposed controller can guarantee arbitrarily small error bound by choosing sufficiently large control gains. Next, a two-step trajectory-optimization-based whole-body motion planning algorithm was proposed while taking omnidirectionality of the OAM and physical constraints including collision avoidance into account. We composed the two offline and online planning phases to formulate a numerically stable optimization problem. 

The proposed control and whole-body motion planning framework is validated in hardware experiments. In the first experiment, we compared the proposed controller with a nonlinear PID controller. 
The proposed controller outperformed the counterpart by showing better regulation performance in the presence of external disturbance by the manipulator. The second and third experiments demonstrated effectiveness of the proposed framework in precise manipulation where the OAM conducted grasping-and-pulling of an object on 1) the ground and 2) a table. We accomplished precise manipulation even in the presence of external disturbance involving ground effect while whole-body motion was exploited in realizing abidance to physical constraints and task completion. During the experiments of precise manipulation, the OAM maintained stable flight around the pitch angle over $90^\circ$ and even $180^\circ$, showing mobile manipulation in arbitrary base pose.

As a future work, we aim to explore collaborative transportation where multiple OAMs transport a common object while each agent exploiting the enlarged workspace. Compared to aerial manipulators based on conventional underactuated multirotors, the additional advantage of the OAMs' enlarged workspace would enhance manipulability of the transporting object, particularly in confined environments.
\blue{
In structured environments, such as warehouses and automated manufacturing facilities, obstacle locations are typically known in advance. However, we also consider future research directions involving dynamic and unknown obstacles, which are common in settings like disaster response or human–robot coexistence.
Our framework may be combined with high-confidence motion prediction for moving agents to address dynamic obstacles.
Furthermore, for unknown environments with static obstacles, the detection of a potential collision using onboard sensors such as cameras can serve as a trigger to re-initiate the planner. This allows the robot to adaptively update its motion plan while preserving whole-body manipulation capabilities and ensuring collision avoidance.
In addition to these, when manipulating heavy objects, it can be beneficial to consider the effects of gravity and the feasible wrench space. Thus, we also consider real-time adaptation of the OAM’s configuration to simultaneously minimize the adverse effects caused by the payload's weight at the end-effector.
}


{
\section{Appendix}

\subsection{Definition of $A$ in subsection \ref{subsec:control allocation}} \label{appendix-A}
We divide the matrix $\bm{A}$ into two submatrices $\bm{A}_1, \bm{A}_2 \in \mathbb{R}^{6\times 6}$ as $\bm{A} = [\begin{matrix}\bm{A}_1 & \bm{A}_2 \end{matrix}]$ where $\bm{A}_1, \bm{A}_2$ are defined as
\begin{equation*}
\bm{A}_1 = \begin{bmatrix}
    0 & -c\frac{\pi}{3} & 0 & -1 & 0 & -c\frac{\pi}{3} \\
    0 &  s\frac{\pi}{3} & 0 & 0  & 0 & -s\frac{\pi}{3} \\
    1 & 0               & 1 & 0  & 1 & 0 \\\
    -Lc\frac{\pi}{3} & k_f c\frac{\pi}{3} & -L & -k_f & -L c\frac{\pi}{3} & k_f c \frac{\pi}{3} \\
    Ls\frac{\pi}{3} & -k_f s\frac{\pi}{3} & 0 & 0 & -L s\frac{\pi}{3} & k_f s \frac{\pi}{3} \\
    -k_f & -L & k_f & -L & -k_f & -L 
\end{bmatrix}
\end{equation*}    
\begin{equation*}
\bm{A}_2 = \begin{bmatrix}
 0 & c\frac{\pi}{3}  & 0 & 1 & 0 & c\frac{\pi}{3} \\
 0 & -s\frac{\pi}{3} & 0 & 0 & 0 & s\frac{\pi}{3} \\
 1 & 0               & 1 & 0 & 1 & 0 \\
Lc\frac{\pi}{3} & k_f c\frac{\pi}{3} & L & -k_f & L c\frac{\pi}{3} & k_f c \frac{\pi}{3} \\
-Ls\frac{\pi}{3} & -k_f s\frac{\pi}{3} & 0 & 0 & L s\frac{\pi}{3} & k_f s \frac{\pi}{3} \\
k_f & -L & -k_f & -L & k_f & -L 
\end{bmatrix}.
\end{equation*}    
Here, $L, k_f \in \mathbb{R}_{>0}$ are constant values denoting half of the maximal length between any two rotors and the thrust to torque ratio of a single rotor, respectively. In our experiment, we set $L = 0.018$ \si{m} and $k_f = 0.015$ \si{m}.

\subsection{Proof of Lemma \ref{lemma 3}} \label{appendix: proof of lemma 3}
\begin{proof}
Using the Comparison Lemma \cite[Lemma 3.4]{khalil2002nonlinear}, $s(t) \leq \bar{s}(t)$ where $\bar{s}(t)$ satisfies $\dot{\bar{s}} = -\alpha(\bar{s}) + \alpha(c_s)$ and $\bar{s}(0) = s(0)$. Considering a positive definite function $H = \frac{1}{2}(\bar{s}-c_s)^2$ for $\bar{s}-c_s$, its time-derivative is computed as $\dot{H} = -(\bar{s} - c_s)(\alpha(\bar{s}) - \alpha(c_s))$, which is negative definite. Thus, using \cite[Theorem 4.9]{khalil2002nonlinear}, there exists $\beta(\cdot,\cdot) \in \mathcal{KL}$ such that $\bar{s}(t) \leq \beta(\bar{s}(0)-c_s,t) + c_s$ $\forall t\geq 0$. Since $s(t) \geq \bar{s}(t)$ $\forall t$ and $\bar{s}(0) = s(0)$, this finishes the proof.    
\end{proof}

\subsection{Proof of Theorem \ref{theorem 1}} \label{appendix: proof of theorem 1}
\begin{proof}
    By using (\ref{eq: Ntilde t}), Lemma 2 and Young's inequality for obtaining $\bm{e}_p^\top \bm{e}_{t1} \leq \frac{1}{2} \lVert \bm{e}_p \rVert^2 + \frac{1}{2} \lVert \bm{e}_{t1} \rVert^2$ and $\bm{e}^\top_{t2} \tilde{\bm{N}}_t \leq \lambda_{m}(\bm{K}_{ti}) \lVert \bm{e}_{t2} \rVert^2 + \frac{1}{4 \lambda_{m}(\bm{K}_{ti})} \lVert \tilde{\bm{N}}_t \rVert^2$, $\dot{V}_t$ can be further arranged as
    \begin{equation} \label{eq: Vdot t arranged}
        \dot{V}_t \leq -\left(\eta^*_t - \frac{\mu_t^2}{4 \lambda_{m}(\bm{K}_{ti})} \right) \lVert \bm{e}_t \rVert^2 + \sum^n_{i=1}\cfrac{\Gamma_{t,i}}{\Theta_{t,i}} c.
    \end{equation}

    Meanwhile, by Lemma 1, $V_t$ satisfies the lower and upper bounds of 
    \begin{equation} \label{eq: Vt lower and upper bound}
    \eta_t \lVert \bm{e}_t \rVert^2 \leq V_t(t) \leq \bar{\eta}_t(\lVert \bm{e}_t \rVert) + \sigma_t
    \end{equation}
    where 
    \begin{equation*}
    \begin{gathered}
        \sigma_t = \sum \frac{\Gamma_{t,i}}{\Theta_{t,i}} \text{ln}2, \quad \eta_t = \min \{\frac{1}{2}m,\frac{1}{2} \} \\
        \bar{\eta}_t(\lVert e_t \rVert) = \max \{\frac{1}{2}m,\frac{1}{2} \} \lVert \bm{e}_t \rVert^2 + \sum^n_{i=1}(\Gamma_{t,i} + \lVert N_{td,i} \rVert_\infty) \lvert e_{t1,i} \rvert.
    \end{gathered}
    \end{equation*}
    Using (\ref{eq: Vt lower and upper bound}) and the fact that $(\alpha(a + b))^2 \leq (\alpha(2 a))^2 + (\alpha(2 b))^2$ for any $\alpha(\cdot) \in \mathcal{K}_\infty$ and $a,b\in\mathbb{R}$, and $\bar{\eta}^{-1}_t(\cdot) \in \mathcal{K}_\infty$, i.e. the inverse of $\bar{\eta}_t(\cdot) \in \mathcal{K}_\infty$, 
    \begin{equation*}
    \begin{aligned}
        \lVert \bm{e}_t \rVert^2 &\geq 
        \begin{cases} 
            \{\bar{\eta}^{-1}_t(V_t - \sigma_t) \}^2 & \text{if }V_t \geq \sigma_t \\ 
             0 & \text{if }V_t < \sigma_t 
        \end{cases} \\
        &\geq
        \begin{cases} 
            \{\bar{\eta}^{-1}_t(V_t/2) \}^2 - \{\bar{\eta}^{-1}_t(\sigma_t) \}^2 & \text{if }V_t \geq \sigma_t \\ 
             0 & \text{if }V_t < \sigma_t 
        \end{cases}.
    \end{aligned}
    \end{equation*}
    Then, by applying the control gain condition and the above inequality, 
    \begin{equation*}
        \dot{V}_t \leq -\Omega_t(V_t) + \Omega_t(\Xi_t)
    \end{equation*}
    where 
    \begin{equation*}
    \begin{aligned}
        \Omega_t(V_t) &= \cfrac{\eta^*_t}{2} (\bar{\eta}_t^{-1}(V_t/2))^2 \in \mathcal{K}_\infty \\
        C_t(\sigma_t) &= \cfrac{\eta^*_t}{2} (\bar{\eta}_t^{-1}(\sigma_t))^2 + \cfrac{c}{\text{ln}2} \sigma_t
    \end{aligned}    
    \end{equation*}
    and $\Xi_t = \Omega_t^{-1}(C_t(\sigma_t))$. By applying Lemma 3, the following upper bound for $V_t$ can be obtained, and that for $\lVert \bm{e}_t \rVert$ also naturally follows from (\ref{eq: Vt lower and upper bound}):
    \begin{equation*}
    \begin{gathered}
        V_t \leq \beta_t(\lvert V_t(0) - \Xi_t \rvert,t) + \Xi_t, \\
        \lVert \bm{e}_t \rVert \leq \sqrt{\cfrac{1}{\eta_t} \beta_t(\lvert V_t(0) - \Xi_t \rvert,t) + \cfrac{\Xi_t}{\eta_t}}
    \end{gathered}
    \end{equation*}
    where $\beta_t(\cdot,\cdot) \in \mathcal{KL}$. Since the ultimate bound $\sqrt{\frac{\Xi_t}{\eta_t}}$ can be made arbitrarily small by taking sufficiently large $\Theta_t$, the proof is complete.
\end{proof}

\subsection{Proof of Lemma \ref{lemma 4}} \label{appendix: proof of lemma 4}
\begin{proof}
    Finite $\Gamma_{r,i}$ exists by the Assumption 1. Since $\Psi(t) \leq \psi < 2$ for $t \in [t_1,t_2]$, $\frac{1}{2} \lVert \bm{e}_R \rVert^2 \leq \Psi(t) \leq \frac{1}{2-\psi} \lVert \bm{e}_R \rVert^2$ holds for all $t \in [t_1,t_2]$ \cite{lee2010geometric}. Then, similar to the result for $V_t$ in (\ref{eq: Vt lower and upper bound}), the lower and upper bounds of $V_r$ can be computed as
    \begin{equation} \label{eq: Vr lower and upper bound}
        \eta_r \lVert \bm{e}_r \rVert^2 \leq V_r(t) \leq \bar{\eta}_r(\lVert \bm{e}_r \rVert) + \sigma_r
    \end{equation}
    where $\eta_r, \sigma_r > 0$ and $\bar{\eta}_r(\cdot) \in \mathcal{K}_\infty$. Here, we use the fact that $Q_r$ has lower and upper bounds similar to the result of Lemma 1 for $Q_t$.
    Similar to (\ref{eq: Vdot t arranged}), $\dot{V}_r$ is upper-bounded by
    \begin{equation*}
        \dot{V}_r \leq -\left(\eta^*_r - \frac{\mu_r^2(\lVert \bm{e}_r \rVert)}{4 \lambda_m(\bm{K}_{ri})} \right) \lVert \bm{e}_r \rVert^2 + \sum \frac{\Gamma_{r,i}}{\Theta_{r,i}} c
    \end{equation*}
    where we use $\dot{\Psi} = \bm{e}_\omega^\top \bm{e}_R$. Then, by following the same procedure in Theorem 1, we can show that $\dot{V}_r \leq -\Omega_r(V_r) + \Omega_r(\Xi_r)$ where $\Xi_r = \Omega_r^{-1}(C_r(\sigma_r))$ and 
    \begin{equation} \label{eq: Omega_r and C_r}
    \begin{aligned}
        \Omega_r(V_r) &= \cfrac{\eta^*_r}{2} (\bar{\eta}_r^{-1}(V_r/2))^2 \in \mathcal{K}_\infty \\
        C_r(\sigma_r) &= \cfrac{\eta^*_r}{2} (\bar{\eta}_r^{-1}(\sigma_r))^2 + \cfrac{c}{\text{ln}2} \sigma_r.
    \end{aligned}    
    \end{equation}
    The definition of $\beta_r(\cdot,\cdot) \in \mathcal{KL}$ is derived when applying Lemma 3 to (\ref{eq: Vr omega bound}).
\end{proof}

\subsection{Proof of Theorem \ref{theorem 2}} \label{appendix: proof of theorem 2}
\begin{proof}
    Let the maximal time interval for $V_r(t) \leq \psi$ is given by $t \in [0,\delta_t]$. Thanks to the initial condition (\ref{eq: initial condition - rotation}) and smoothness of $V_r(t)$, there always exists $\delta_t > 0$. Now, assume that $\delta_t$ is finite. Then, since $[0,\delta_t]$ is the maximum time interval, $V_r(\delta_t) = \psi$ and $\dot{V}_r(\delta_t) \geq 0$. However, from Lemma 4, during the time interval $[0,\delta_t]$, the following holds:
    \begin{equation*}
        \dot{V}_r(t) \leq -\Omega_r(V_r(t)) + \Omega_r(\Xi_r).
    \end{equation*}
    Then, 
    \begin{equation*}
    \begin{aligned}
        \dot{V}_r(\delta_t) &\leq -\Omega_r(V_r(\delta_t)) + \Omega_r(\Xi_r) \\
        &= -\Omega_r(\psi) + \Omega_r(\Xi_r) < 0
    \end{aligned}
    \end{equation*}
    where we use $\Xi_r < \psi$. However, this contradicts to $\dot{V}_r(\delta_t) > 0$; thus, $V_r(t) \leq \psi$ holds for all $t \in [0,\infty)$.

    From this property of $V_r(t) \leq \psi$ $\forall t \in [0,\infty)$ and the fact that $\Psi \leq V_r$, we can apply Lemma 4 again for the infinite time interval $[0,\infty)$. Then, using Lemma 3 and (\ref{eq: Vr lower and upper bound}), 
    \begin{subequations}
    \begin{gather}
        \begin{split} \label{eq: Vr beta}
        V_r \leq \beta_r(\lvert V_r(t_0) - \Xi_r \rvert,t-t_0) + \Xi_r 
        \end{split}\\
        \begin{split} \label{eq: er beta}
        \lVert \bm{e}_r \rVert \leq \sqrt{\frac{1}{\eta_r} \beta_r(\lvert V_r(t_0) - \Xi_r \rvert, t-t_0) + \frac{\Xi_r}{\eta_r}}    
        \end{split}
    \end{gather}
    \end{subequations}
    where the definitions of $\beta_r(\cdot,\cdot)$ and $\Xi_r$ are in Lemma 4. 

    The proof is finished by using (\ref{eq: er beta}) which holds $\forall t \in [0,\infty)$ and the fact that the ultimate bound $\sqrt{\frac{\Xi_r}{\eta_r}}$ can be made arbitrarily small by taking sufficiently large $\Theta_r$ recalling that $\Xi_r = \Omega_r^{-1}(C_r(\sigma_r))$ and $\sigma_r = \sum \frac{\Gamma_{r,i}}{\Theta_{r,i}} \text{ln}2$, where the definitions of $\Omega_r, C_r$ can be found in (\ref{eq: Omega_r and C_r}).
\end{proof}
}

\subsection{Definition of ${}^E f_p$ and ${}^E f_R$ } \label{appendix: def of offline kinematics}
The kinematics of translational and rotational motion represented in the continuous time domain are as follows:
\begin{equation*}
    \begin{aligned}
    \frac{d}{dt}{}^E\bm{p}_d &= {}^E\bm{v}_d, &\frac{d}{dt}{}^E\bm{v}_d &= {}^E\dot{\bm{v}}_d, & \frac{d}{dt}{}^E\dot{\bm{v}}_d &= {}^E\ddot{\bm{v}}_d\\
    \frac{d}{dt}{}^E\bm{R}_d &= {}^E\bm{R}_d {}^E\bm{\omega}_d^{\wedge}, &\frac{d}{dt}{}^E\bm{\omega}_d &= {}^E\dot{\bm{\omega}}_d, &\frac{d}{dt}{}^E\dot{\bm{\omega}}_d &= {}^E\ddot{\bm{\omega}}_d  
    \end{aligned}
\end{equation*}
We numerically integrate these differential equations using the Runge-Kutta $4^{th}$-order method in the offline planning phase to reduce numerical errors. More specifically, for given $^E\bm{x}_p =[{}^E\bm{p}_d;{}^E\bm{v}_d;{}^E\dot{\bm{v}}_d]$, $^E\bm{x}_R=({}^E\bm{R}_d,{}^E\bm{\omega}_d,{}^E\dot{\bm{\omega}}_d)$, ${}^E\ddot{\bm{v}}_d$ and ${}^E\ddot{\bm{\omega}}_d$, the kinematic models ${}^E\bm{f}_p$ and ${}^E\bm{f}_R$ in the discrete time domain can be found below: 
\begin{equation*}
    \begin{gathered}
    \bar{\bm{v}}_1 = \bm{v}_d + 0.5 \Delta t \dot{\bm{v}}_d, \quad\dot{\bar{\bm{v}}}_1 = \dot{\bm{v}}_d + 0.5 \Delta t \ddot{\bm{v}}_d\\
    \bar{\bm{v}}_2 = \bm{v}_d + 0.5 \Delta t \dot{\bar{\bm{v}}}_1, \quad\dot{\bar{\bm{v}}}_2 = \dot{\bm{v}}_d + 0.5 \Delta t \ddot{\bm{v}}_d\\
    \bar{\bm{v}}_3 = \bm{v}_d + \Delta t \dot{\bar{\bm{v}}}_2, \quad\dot{\bar{\bm{v}}}_3 = \dot{\bm{v}}_d +\Delta t \ddot{\bm{v}}_d
    \end{gathered}
\end{equation*}
\begin{equation} \label{planner-fpE}
    {}^E\bm{f}_p =
    \begin{bmatrix}
        {}^E\bm{p}_d + \Delta t /6 ( \bm{v}_d+2\bar{\bm{v}}_1+2\bar{\bm{v}}_2+\bar{\bm{v}}_3 )\\
        {}^E\bm{v}_d +  \Delta t /6 ( \dot{\bm{v}}_d+2 \dot{\bar{\bm{v}}}_1+2\dot{\bar{\bm{v}}}_2+\dot{\bar{\bm{v}}}_3 )\\
        {}^E\dot{\bm{v}}_d +  \Delta t \ddot{\bm{v}}_d
    \end{bmatrix}
\end{equation}
\begin{equation*}
    \begin{gathered}
    \bar{\bm{\omega}}_1 = \bm{\omega}_d + 0.5 \Delta t \dot{\bm{\omega}}_d, \quad\dot{\bar{\bm{\omega}}}_1 = \dot{\bm{\omega}}_d + 0.5 \Delta t \ddot{\bm{\omega}}_d\\
    \bar{\bm{\omega}}_2 = \bm{\omega}_d + 0.5 \Delta t \dot{\bar{\bm{\omega}}}_1, \quad\dot{\bar{\bm{\omega}}}_2 = \dot{\bm{\omega}}_d + 0.5 \Delta t \ddot{\bm{\omega}}_d\\
    \bar{\bm{\omega}}_3 = \bm{\omega}_d + \Delta t \dot{\bar{\bm{\omega}}}_2, \quad\dot{\bar{\bm{\omega}}}_3 = \dot{\bm{\omega}}_d + \Delta t \ddot{\bm{\omega}}_d
    \end{gathered}
\end{equation*}

\begin{equation} \label{planner-fRE}
{}^E\bm{f}_R = 
    \begin{bmatrix}
        {}^E\bm{R}_d\textbf{exp}( \Delta t /6 ( \bm{\omega}_d+2\bar{\bm{\omega}}_1+2\bar{\bm{\omega}}_2+\bar{\bm{\omega}}_3 )^\wedge)\\
        {}^E\bm{\omega}_d +  \Delta t /6 ( \dot{\bm{\omega}}_d+2 \dot{\bar{\bm{\omega}}}_1+2\dot{\bar{\bm{\omega}}}_2+\dot{\bar{\bm{\omega}}}_3 )\\
        {}^E\dot{\bm{\omega}}_d +  \Delta t \ddot{\bm{\omega}}_d
    \end{bmatrix}
\end{equation}
where $\textbf{exp}(\cdot)$ is a matrix exponential. 
We omitted the time index $k$ and the arguments of ${}^E\bm{f}_p$ and ${}^E\bm{f}_R$ for simplicity. 

\subsection{Definition of $f_x$} \label{appendix: def of online kinematics}
The kinematics of OAM in the continuous time domain are represented as the following:
\begin{equation*}
    \begin{gathered}
        \frac{d}{dt}\bm{p}_d = \bm{v}_d, \quad \frac{d}{dt}\bm{R}_d = \bm{\omega}_d, \quad \frac{d}{dt}{\bm{\theta}}_d = \dot{\bm{\theta}}_d\\
    \end{gathered}
\end{equation*}
For fast computation, we numerically integrate the above differential equations using the Euler $1^{st}$-order method. In other words, for given 
$\bm{x}_d(k) = (\bm{p}_d(k), \bm{R}_d(k), \bm{\theta}_d(k))$ and $\bm{u}_k = (\bm{v}_d(k), \bm{\omega}_d(k), \dot{\bm{\theta}}_d(k))$,
we consider the following as the definition of $\bm{f}_x(\bm{x}_d(k),\bm{u}_d(k))$:
\begin{equation} \label{planner-fx}
    \bm{f}_x = 
    \begin{bmatrix}
        \bm{p}_d(k) + \Delta t \bm{v}_d(k) \\ 
        \bm{R}_d(k)\textbf{exp}( \Delta t (\bm{\omega}_d(k))^{\wedge} ) \\ 
        \bm{\theta}_d(k) + \Delta t \dot{\bm{\theta}}_d(k)
    \end{bmatrix}.
\end{equation}

\end{document}